\pdfoutput=1

\documentclass[11pt]{article}

\usepackage{acl}

\usepackage{times}
\usepackage{latexsym}

\usepackage[T1]{fontenc}

\usepackage[utf8]{inputenc}

\usepackage{microtype}

\usepackage{url}
\usepackage{color}
\usepackage{subcaption,graphicx}
\usepackage{float}

\usepackage{multirow}
\usepackage{ulem}
\usepackage{enumitem}
\usepackage{array}
\usepackage{cleveref}
\usepackage{booktabs}

\newcounter{takeaway}

\title{On Linearizing Structured Data in Encoder-Decoder Language Models:\\ Insights from Text-to-SQL}

\author{ Yutong Shao \and Ndapa Nakashole \\
        Computer Science and Engineering \\
        University of California, San Diego \\
        La Jolla, CA 92093 \\
        \texttt{\{yshao, nnakashole\}@eng.ucsd.edu}}

\begin{document}
\maketitle

\begin{abstract}
Structured data, prevalent in  tables, databases, and knowledge graphs, poses a significant challenge in its representation. With the advent of large language models (LLMs), there has been a shift towards linearization-based methods, which process structured data as sequential token streams, diverging from  approaches that explicitly model structure, often as a  graph.  Crucially, there remains a gap in our understanding of how these  linearization-based methods handle structured data, which is inherently  non-linear.
This work  investigates   the linear handling of structured data  in encoder-decoder language models, specifically  T5.  Our findings  reveal the model's ability to mimic human-designed processes such as  schema linking and syntax prediction, indicating a deep, meaningful learning of structure beyond simple token sequencing. We also uncover insights into the model's internal mechanisms, including the ego-centric nature of structure node encodings and the potential for model compression due to modality fusion  redundancy. Overall, this work sheds light on the inner workings of linearization-based methods and could potentially provide guidance for future research.

\end{abstract}

\section{Introduction}


\textbf{Motivation.} Natural Language Interfaces (NLIs) to computer systems allow the use  of  everyday language to interact with computer systems, thus  lowering technical barriers to advanced computing functionality. Early systems such as SHRDLU~\cite{winograd1971procedures} and LUNAR~\cite{woods1973progress} saw  limited  success due to the limited language processing capabilities of computer systems at the time.  After years of steady progress,  the strong language processing capabilities  of large language models (LLMs) have led to renewed interest in NLIs such as the widely used ChatGPT~\cite{DBLP:conf/nips/BrownMRSKDNSSAA20}.
Systems such as ChatGPT  already serve as robust NLIs. However, a critical challenge remains in applying the underlying models to specialized and personalized real-world scenarios. The challenge stems from the need for the model to handle ``backend data'' commonly stored in structured formats such as proprietary databases, knowledge graphs, or dialog states, including intents, slots, and values. We refer to this task as \textit{structured data representation} (SDR)\cite{DBLP:journals/debu/Shao0N22}. 

In this study, our primary focus is on a  representative SDR task: text-to-SQL parsing~\cite{zelle1996learning, zettlemoyer2012learning}. This task automatically maps natural language queries into SQL commands, thus eliminating or at least reducing the need for programming knowledge. For such a system to be broadly applicable, it must be capable of adapting to new databases, encoding these databases together with user queries, and predicting the corresponding SQL queries.

\paragraph{The Rise of Linearization-based Methods.} 
Recent approaches to text-to-SQL parsing and other SDR tasks fall into two main categories: \textit{linearization-based} and \textit{structure-based} methods~\cite{bridge:Lin2020BridgingTA,picard:scholak2021picardpi,uskg:xie2022unifiedskgua}. Structure-based approaches explicitly utilize the inherent  structure in the data, often representing it with a graph~\cite{schemagnn:Bogin2019RepresentingSS,ratsql:dblp:conf/acl/wangslpr20,lgesql:Cao2021LGESQLLG,s2sql:Hui2022S2SQLIS}. In contrast, linearization-based methods treat  structured data as a token sequence, processed similarly to natural language sentences. The latter have gained traction due to their compatibility with LLMs, which have demonstrated impressive performance across various NLP benchmarks. 

\paragraph{Open Problems and Our Contributions.}
SDR tasks like text-to-SQL remain a challenging problem for LLMs, as they are not completely ``solved'' by current models~\cite{Li2023CanLA}. 
Towards shedding light on ways forward, our main contribution is a detailed exploration of the inner workings of a former state-of-the-art (SOTA) text-to-SQL parser with T5 backbone~\cite{picard:scholak2021picardpi,uskg:xie2022unifiedskgua}.
Our analytical approaches include probing classifiers~\cite{DBLP:conf/emnlp/Kohn15,DBLP:conf/emnlp/GuptaBBP15,DBLP:conf/emnlp/ShiPK16,conneau2018what}, as well as   techniques that directly manipulate model intermediates, leveraging the recent causal tracing method~\cite{DBLP:conf/acl/FinlaysonMGSLB20, rome:meng2022locatingae}.
We find that despite their simplicity, linearization-based methods can effectively represent structured data. Specifically, we show that the prefix-tuned T5 model preserves low-level textual details and enhances understanding of node relationships in structured data. We also show the ego-centric nature of structure node encodings, which primarily contain information relevant to the node itself. Additionally, we uncover a duplicative robustness in modality fusion, indicating potential avenues for model compression. Our study also reveals the model's internal working pipeline, which aligns with human-designed processes like schema linking, syntax prediction, and node selection, suggesting meaningful learning rather than reliance on spurious correlations. The attention mechanism's role in modality fusion and the distinctive functionalities of different layer ranges in the decoder are also revealed. Overall, our research contributes to our understanding of structured data representation in encoder-decoder LLMs.

We opted not to analyze extremely large LMs such as GPT-4, due to the high computational cost of our analytical methods and the opaque nature of their intermediate states. 
However, given the competitive performance of the model we study, coupled with its sequential input and autoregressive output which are analogous to LLMs, we believe that our findings are general and applicable to models in the same category.



\section{Related Work}


\paragraph{Structured Data Representation for Text-to-SQL.}
From prior work, structure-based methods include SchemaGNN~\cite{schemagnn:Bogin2019RepresentingSS}, RAT-SQL~\cite{ratsql:dblp:conf/acl/wangslpr20}, LGE-SQL~\cite{lgesql:Cao2021LGESQLLG} and S$^2$SQL~\cite{s2sql:Hui2022S2SQLIS}.
Linearization-based methods have been widely studied, including BRIDGE~\cite{bridge:Lin2020BridgingTA} and Picard~\cite{picard:scholak2021picardpi}. USKG~\cite{uskg:xie2022unifiedskgua}, which proposes a unified linearization method for all SDR tasks, also falls under the category. 
Recently, LLMs such as those behind ChatGPT also demonstrated strong performance as a structure linearization-based text-to-SQL method~\cite{Li2023CanLA}.


\paragraph{Model Behavior Analysis and Interpretation.} Previous work include gradient-based methods which check the importance of input features based on their gradient, such as saliency maps~\cite{saliency:Simonyan2013DeepIC}. For models reliant on attention mechanisms~\cite{ DBLP:conf/blackboxnlp/ClarkKLM19}, analytical methods exist to examine the significance of individual input units by evaluating attention weights. However, such analyses have faced skepticism in other works~\cite{is-attention-int:serrano-smith-2019-attention}.
An alternative line of work is the \textit{probing classifier} approach, in which  classifiers are trained on a model's intermediate representations to determine the existence of specific information~\cite{DBLP:conf/emnlp/Kohn15,DBLP:conf/emnlp/GuptaBBP15,DBLP:conf/emnlp/ShiPK16, DBLP:conf/repeval/EttingerER16, probing:Adi2016FinegrainedAO,DBLP:conf/naacl/Liu0BPS19,DBLP:conf/iclr/TenneyXCWPMKDBD19,DBLP:conf/emnlp/HewittL19,DBLP:conf/emnlp/VoitaT20,DBLP:conf/emnlp/ZhuR20,DBLP:conf/acl/PimentelVMZWC20,DBLP:conf/eacl/RavichanderBH21}. Though flexible and adaptable to trace various information, probing tests may present challenges in the interpretation or comparison of results~\cite{probing-the-probing:ravichander-etal-2021-probing,belinkov-2022-probing}.
A recent line of work  involves causal analysis, wherein researchers manipulate the target information within the input and restore intermediate outcomes to a clean state to verify the presence of information~\cite{DBLP:conf/acl/FinlaysonMGSLB20,rome:meng2022locatingae}.  
Our analysis framework builds upon previous probing and causal analysis methods. However, we have adapted and incorporated additional methods specifically to gain insights into tasks related to structured data representation.

\begin{figure*}[t]
\centering
\includegraphics[width=0.65\linewidth]{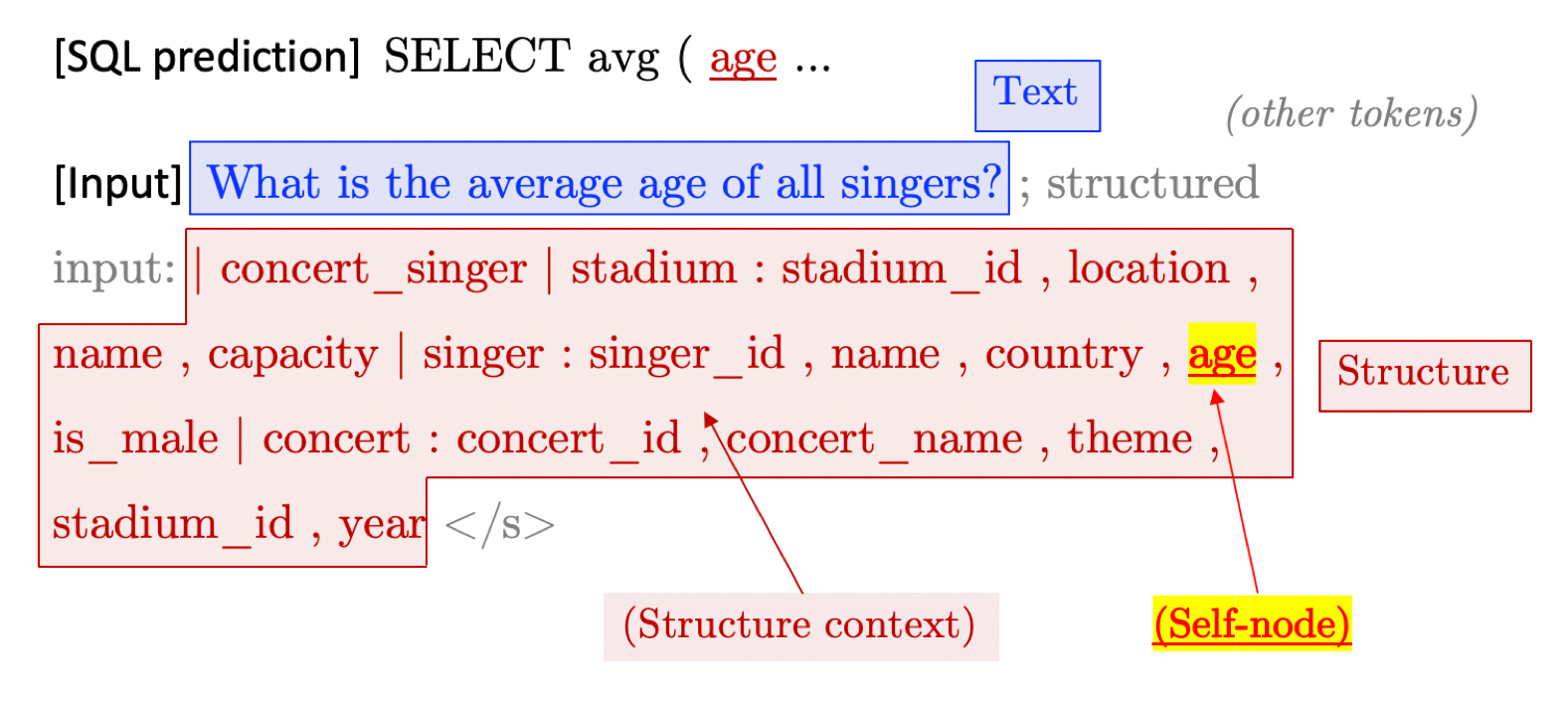}
\caption{The input to the text-to-SQL parser consists of the query in natural language text (blue), and the relevant structured data (red), other tokens (gray). ``self-node,'' refers to the input tokens corresponding to the expected output node where a node  refers to both column and table names, and ``structure-context,'' represents all the structured input tokens excluding the 
self-node. The output is the predicted SQL query (top). }
\label{fig:input-sections}
\end{figure*}
\section{Preliminaries}
The model we investigate in this work is T5-large,\footnote{The prior SOTA method for Spider is Picard + T5-3B~\cite{picard:scholak2021picardpi}, in which Picard is a post-hoc decoding algorithm without modifying the model. In the scope of our experiments,
we use T5-large instead, as our behavior tests involve significant computational expenses.} with prefix-tuning on the Spider dataset~\cite{yu2018spider,uskg:xie2022unifiedskgua}.
The model is a standard encoder-decoder Transformer architecture.\footnote{As our focus is on the interpretation of the inner workings of the model, we do not use any post-processing heuristic on the model output, like Picard.} 
The encoder has two modules per layer: self-attention and  linear multilayer perceptron  (MLP). A decoder layer has three modules: self-attention, cross-attention, and MLP.
To work as a text-to-SQL parser, the model takes the concatenation of user query text and the linear form of a database (DB) structure as input and yields the SQL token sequence as output. 
The model has been trained with \textit{prefix-tuning}~\cite{DBLP:conf/acl/LiL20}. 
Each attention module within the model is associated with $10$ extra key-value entries from the prefix.

\paragraph{Terminology.} For clarity of exposition, we introduce the term "structure nodes," or simply "nodes," which collectively refer to both columns and tables.
We refer to ``layer ranges'' in the encoder or decoder, grouping them into to low, middle, and high layers. Both the encoder and decoder have 24 layers. In our discussion, ``low layers'' refer to layer 0-11, ``middle'' to 6-17 and ``high'' to 12-23.
In addition, we mention ``input sections,'' such as text, structure, and prefix. 
For a structure node, we also have the input section of ``self-node,'' indicating the input tokens
of the anticipated output node, and ``structure-context,'' representing all the structured input tokens excluding the 
self-node. Figure~\ref{fig:input-sections} illustrates the format of the  input  to the text-to-SQL parser.

\section{Research Questions}



\paragraph{Preliminary Intuition Open Questions.} The model's internal mechanics can be intuitively understood as follows: The encoder produces contextualized encodings by fusing both textual and structural input elements. These combined encodings then provide a comprehensive basis upon which the decoder constructs the SQL. However,  despite this preliminary understanding, certain aspects remain unclear. Key questions that arise include: 


\textit{(Q1) What specific information is passed from the encoder to the decoder via the text and structure token encodings?} Addressing this allows us to delineate the functions of the encoder and decoder, further aiding in model interpretation.

\textit{(Q2) Which parts of the model store the important information?} Here ``parts'' refers to model intermediates within different modules, input sections or layer ranges. This helps in detecting the possible bottlenecks and presents opportunities for model compression on the less important parts.


\textit{(Q3) How do the attention modules handle modality fusion?}
In our exploration of structured data representation in text-only models, we aim to understand how modality fusion is performed. It is expected to occur within the attention modules, as they are the only mechanism enabling different tokens to communicate.

\textit{(Q4) What is the internal working pipeline of the model?} We aim to check if the thought process of the model mirrors human-designed pipelines, including steps such as schema-linking, syntax prediction, and node selection~\cite{Gu2023InterleavingPL,dinsql:Pourreza2023DINSQLDI}. Additionally, we seek to verify whether the model obtains meaningful knowledge through text-to-SQL training, or it is mainly fitting to spurious correlations.


\section{Probing Study}
\label{sec:probing}

\subsection{Probing Tasks}
For our first question \textbf{Q1}, our initial investigation focused on the information retained by the encoding vectors. For this purpose, we conducted two probing tasks as described below.

\paragraph{Node Name Reconstruction (NR).} 
In this task, we examine the ability of the encoder to retain essential, low-level information about a node by attempting to reconstruct its surface-form name.
Due to the tokenization process of T5, a node can be tokenized into multiple sub-tokens.
We keep all sub-token encodings as a sequence and pass them to a ``probe decoder,'' which has the same architecture as T5-decoder but is initialized randomly. 
The probe decoder is trained to reconstruct the node name autoregressively.

\paragraph{Link Prediction (LP).}
Unlike node name reconstruction, this task 
assesses the ability of the model to capture higher-order structural information in its encodings.
Using the encodings of a pair of nodes,\footnote{In this study, a question token is also considered a node.} we train a probing classifier to predict the connection between them. The relation definitions follow RAT-SQL~\cite{ratsql:dblp:conf/acl/wangslpr20}. Examples of these relations include  QT-Exact (Question token and Table name Exact match), 
CC-TableMatch (two Columns belong to the same Table), or null relations like XY-default (a type X node and a type Y node, no special relation).
We pool the sub-token encodings of each node into a single vector. We then construct the input vector of two nodes by concatenating their pooled encodings and their element-wise dot-product, i.e. $[e_1;e_2;e_1*e_2]$ where ``;'' denotes concatenation and ``*'' denotes element-wise dot-product. 
We pass this input vector to a probe classifier, which is either a logistic regressor (LR) or a 2-layer MLP,
to predict the relation between the two input nodes.


\subsection{Probing Results}
For both probing tasks, we utilized the train and dev partitions from the Spider dataset for training and evaluation respectively.
The results are presented in Table~\ref{tbl:probing-NR-LP}. For Node Name Reconstruction (NR), both prefix-tuned and pretrained T5 exhibit very high reconstruction accuracy. 
This indicates that the prefix-tuning process did not undermine the ability of the model to preserve the low-level information.

For Link Prediction (LP), our results showed that the encodings from prefix-tuned T5 model outperformed the pre-trained version, suggesting that the prefix-tuning process of the model enhances its understanding of relations between nodes. 
Interestingly, the pre-trained T5 model also yields high LP accuracy. 
This suggests that even without tuning, pre-trained models have an implicit capability to process structured text.
This observation is consistent with findings from~\cite{probing-the-probing:ravichander-etal-2021-probing}, where a model can learn features that are not aligned with its primary objective.

For comparison, the T5-random version showed significantly lower probing performance compared to either T5-prefix-tuned or pre-trained. This confirms that the high performance are not merely due to overfitting noise in high dimensions.


\stepcounter{takeaway}
 \paragraph{Takeaway~\thetakeaway~(Q1).} Regarding the information contained in the encodings, \textbf{\textit {prefix-tuned T5 manages to preserve low-level textual details and also improves  understanding of node relationships}}.  Surprisingly, \textbf{\textit { pre-trained T5 model also exhibits an intrinsic capacity to handle structured text to some degree.}}


\begin{table}[t]
\centering
\resizebox{0.9\linewidth}{!}{%
\begin{tabular}{|c|c|c|c|}
\hline
\multirow{2}{*}{Model}       & NR  & \multicolumn{2}{c|}{LP} \\\cline{2-4}
                & Exact Match  & LR acc.  & MLP acc. \\\hline
T5-P-tuned   & 0.9649 & 0.8110	& 0.8600 \\\hline
T5-pretrain    & 0.9709 & 0.7929	& 0.8400 \\\hline
T5-random      & 0.4918 & 0.2839	& 0.3102 \\\hline

\end{tabular}}
\vspace{-0.1cm}
\caption{Node Name Reconstruction (NR) and Link Prediction (LP) probing results. ``P-tuned'' represents the T5 model with prefix-tuning.}
\vspace{-0.3cm}
\label{tbl:probing-NR-LP}
\end{table}



\section{Direct Model Manipulation}
Besides the insights from the probing study, a remaining question is whether the model actually \textit{utilizes} the information encoded in the representation.
Since probing experiments did not answer this question, we undertook an exploration of the internal mechanisms of the model by directly manipulating the model intermediates and observing the outcomes.
We employ the idea of \textit{causal tracing}~\cite{DBLP:conf/acl/FinlaysonMGSLB20,rome:meng2022locatingae}, where specific intermediate information is corrupted or reinstated to analyze its influence on the ultimate prediction.
We design our studies in a fine-grained way, computing prediction accuracy at the token-level within a SQL query and categorizing the outcomes based on token types. Token types include \textit{columns}, \textit{tables}, \textit{table aliases}, and \textit{syntax tokens} (including keywords and operators). 
We  focus on columns and syntax tokens, treating columns as being representative  of structure node prediction. The results for rest of the token types are in the appendix.

\subsection{Encoder States Investigation}


We begin by \textit{corrupting} the input embeddings or final encoding vectors of individual tokens or entire input sections. A vector is ``corrupted'' by replacing it with a zero-vector. The underlying intuition is that, when corrupting the embeddings of a token or section, the information is fully removed from the model. As such, the corruption study evaluates the overall usefulness of the corrupted part. When the final encodings are corrupted, only the information stored in these encodings is lost, thus this corruption study checks the actual information stored in these encodings and its importance.

\begin{figure}[t]
\centering
\includegraphics[width=\linewidth]{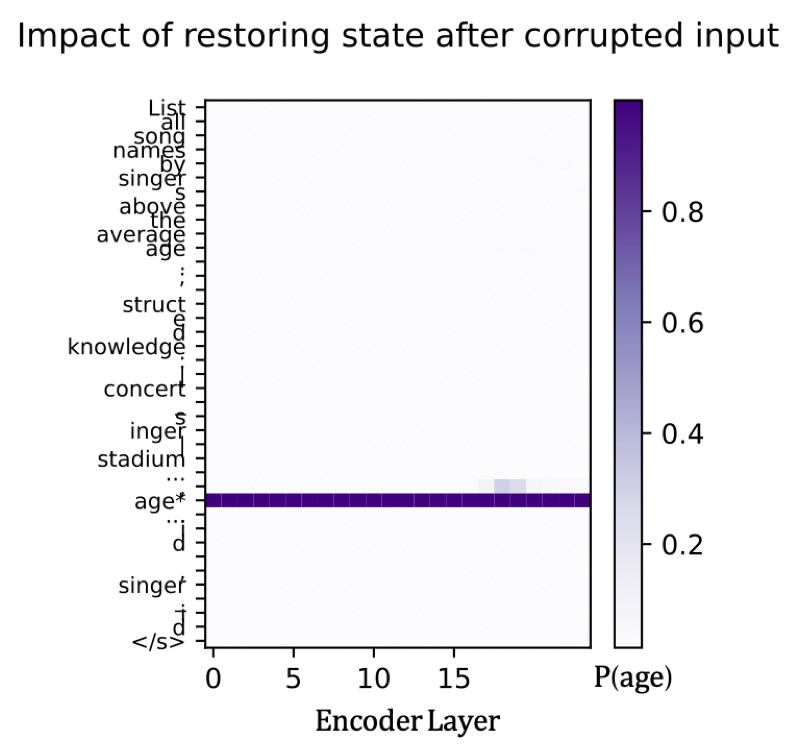}
\vspace{-0.2cm}
\caption{An illustrative sample showing the restoring effect of each encoder intermediate state. The decoder prompt: \texttt{SELECT song\_name FROM singer WHERE ==> age}.
Restoring the self-node hidden state on any layer can recover the correct prediction, while almost all other states do not have such an effect.
More samples are available in Figure~\ref{fig:exp1-appendix}.
} 
\vspace{-0.3cm}
\label{fig:exp1}
\end{figure}

We compute the average confidence on ground truth tokens for samples where the uncorrupted predictions are correct. The results for corrupting entire sections can be found in Table~\ref{tbl:exp2.3+6.0}, on the left.
First, as a sanity check, we confirm that corrupting the embeddings of the entire text section has a substantial impact on the prediction performance of both columns and syntax tokens. For structure embeddings, corrupting column names suffers a larger drop compared to syntax, consistent with our expectation.
For final encodings corruption, we found that when predicting columns, the corruption on merely the self-node is as impactful as the entire structure section. Corruptions on the text part are considerably less effective, and on the structure context it has almost no influence. Shedding light on our question \textbf{Q2} regarding which states hold important information, these results indicate that the self-node encoding vectors are the most important when predicting this node. The structure node encodings are ``ego-centric,'' each storing the information pertinent to that node and not others. 

We further explored the opposite direction, corrupting the text embeddings and \textit{restoring} the intermediate hidden states. 
Following ROME~\cite{rome:meng2022locatingae}, this approach highlights intermediate states with high \textit{restoring effect}, which refers to the increased probability of correct prediction when restoring the state back to the clean version. These states likely capture crucial information and hold notable importance.
We chose to corrupt the embeddings of the text section based on the reasoning that the expected SQL output is entirely specified by the text.

Figure~\ref{fig:exp1} illustrates the restoring effect of all encoder states on a representative sample. We observe that restoring the self-node hidden states at any layer can restore the correct prediction. However, for other tokens, restoring their representations at any layer has minimal impact. This observation reinforces the finding that self-node representations hold greater significance than other tokens when predicting that node.
The ability of a single encoding to restore the correct prediction confirms that the low-level textual information is retained within the encoding and effectively utilized by the decoder  \textbf{(Q1)}.

We also investigated the effects of restoring the final encodings of entire sections with corrupted text embeddings. The findings are presented in Table~\ref{tbl:exp2.3+6.0}, on the right side. We observed the same trend where restoring only the self-node encoding proves to be more effective than restoring the entire text section or the structural context.

\stepcounter{takeaway}
\paragraph{Takeaway~\thetakeaway~(Q1, Q2).} The \textbf{\textit {encodings of structure nodes are predominantly "ego-centric,"}} containing primarily information relevant to themselves with minimal data about other nodes. Consequently, the \textbf{\textit{target node's encodings emerge as the most important}} among all encodings during node prediction.

\begin{table}[t]
\centering
\resizebox{\linewidth}{!}{%
\begin{tabular}{|l|cc|cc||c|}
\hline
\multirow{2}{*}{Section} & \multicolumn{2}{c|}{Embeddings} & \multicolumn{2}{c||}{Final Encodings} & Encoding Restore\\
\cline{2-6}
& Column & Syntax & Column & Syntax & Column \\\hline
Text & 0.2482 & 0.2704 & 0.9115 & 0.4329 & 0.5663 \\\hline
Struct & 0.4084 & 0.8435 & 0.4822 & 0.7056 & 0.8016 \\\hline
Self-node & 0.3916 & - & 0.5239 & - & 0.6801 \\\hline
StructCtx. & 0.8995 & - & 0.9848 & - & 0.1028 \\\hline
all & 0.0083 & 0.0422 & 0.0943 & 0.1458 & 0.9416 \\\hline
\end{tabular}
}
\vspace{-0.1cm}
\caption{The effect of corrupting different sections of the input on columns and syntax token. Values are the average confidence scores of the ground truth over all samples where the clean (uncorrupted) prediction is correct.}
\label{tbl:exp2.3+6.0}
\end{table}

\subsection{Contextual Representations of Structure}

We now turn our attention to \textbf{Q3}, which explores the attention mechanism and the integration of modalities between text queries and structured input.
As can be seen in Table~\ref{tbl:exp2.3+6.0}, when text embeddings are corrupted, even if their final encodings are restored, the prediction accuracy is still compromised ($0.5663$), since the structure nodes fail to access accurate text information. This \textbf{\textit {underscores the important role of text information in the process of encoding structure node, affirming the overall necessity of modality fusion}}.

In the following experiments, we aim to study the inner workings of the attention modules by examining and manipulating them through various methods.

\subsubsection{Attention Corruption Study}

To reveal the inner mechanism of modality fusion, a straightforward first question we propose is: \textit{Where does the fusion primarily take place?} For this, we explore another type of causal study, namely \textit{attention corruption}, in which we deactivate the attention within certain layers and between certain sections by masking the corresponding attention entries.
In detail, there are two corruption schemes, by \textit{``weights''} or by \textit{``logits''}. The ``weights'' setting simply sets the corrupted attention weights to 0, keeping others unchanged. The ``logits'' setting adjusts attention logits to $-\infty$ before the softmax operation, essentially zeroing out the corrupted sections while ensuring a valid distribution.
Both settings are considered in our experiments.

Intuitively,  modality fusion is expected to occur within encoder self-attention or decoder cross-attention.
We evaluated the prediction accuracy of the model for columns and syntax tokens with our above-mentioned attention corruption, targeting the encoder self-attention or the decoder cross-attention, across different sections and within different layer ranges.
Our guiding intuition here is straightforward: for components not engaged in modality fusion, a smaller performance drop is expected.

The results are shown in Table~\cref{tbl:exp5.3+6.1,tbl:exp5.4+6.2}. 
Table~\ref{tbl:exp5.3+6.1} reveals an interesting finding under column ``Columns - Weights'' and on the row ``$S\rightarrow T$'' (structure-to-text\footnote{Th notation of  attention ``from section A to B,''  means the attention matrix entries with tokens from A as query $q$, and tokens from B as keys $k$ and values $v$.}): 
the interference of encoder self-attention from the entire structure section to the text section had a negligible negative impact ($0.9071$), compared to the performance drop in Table~\ref{tbl:exp2.3+6.0} caused by text embeddings corruption ($0.5663$).
Likewise, Table~\ref{tbl:exp5.4+6.2} row ``Text'' reports minimal damage to accuracy ($0.9063$) when the decoder cross-attention to the text was obstructed.

These findings seem to contradict our previous conclusion that the fusion of text and structure information is vital for the task. We propose an explanation for this inconsistency, suggesting that it stems from the \textit{duplicative robustness} of the model. It means the model has learned certain capabilities in multiple locations, encoder and decoder in our case. 
To verify this hypothesis, we experiment with jointly corrupting the encoder self-attention from structure to text, and decoder cross-attention to text, essentially combining the corruption effect of the two experiments above. The results are shown in Table~\ref{tbl:exp5.5+6.3}.
We can observe that 
simultaneously corrupting both leads to a more substantial decline in accuracy. This supports our duplicative robustness hypothesis with regard to modality fusion. Related to \textbf{Q3}, this finding provides novel insights into modality fusion. It underscores the robustness of the model, but also hints at potential opportunities for post-hoc model compression.


\begin{table}[t]

\begin{subtable}{\linewidth}
\centering
\resizebox{\linewidth}{!}{%
\begin{tabular}{|l|cc|cc|}
\hline
\multirow{2}{*}{Corruption part} & \multicolumn{2}{c|}{Columns} & \multicolumn{2}{c|}{Syntax tokens} \\
\cline{2-5}
 & Weights & Logits & Weights & Logits \\\hline
$T\rightarrow S$ & 0.9671 & 0.9543 & 0.9933 & 0.9919 \\\hline
$S\rightarrow T$ & 0.9071 & 0.6879 & 0.9926 & 0.9845 \\\hline
$T\leftrightarrow S$ & 0.8416 & 0.6138 & 0.9878 & 0.9788 \\\hline
all & 0.2101 & 0.1502 & 0.7269 & 0.7338 \\\hline
\end{tabular}
}
\end{subtable}

\caption{Attention corruption study on \textbf{encoder self-attention} across input sections. For input sections, ``T'' means text and ``S'' stands for structure.
``$T\rightarrow S$'' means corrupting the attention weights from text to structure tokens; ``all'' means corrupting the full attention matrix.
On top, ``Weights'' and ``Logits'' represent the attention corruption scheme.
}
\label{tbl:exp5.3+6.1}
\end{table}

\begin{table}[t]



\begin{subtable}{\linewidth}\centering
\resizebox{\linewidth}{!}{%
\begin{tabular}{|l|cc|cc|}
\hline
\multirow{2}{*}{Corruption part} & \multicolumn{2}{c|}{Columns} & \multicolumn{2}{c|}{Syntax tokens} \\
\cline{2-5}
 & Weights & Logits & Weights & Logits \\
\hline
Text & 0.9063 & 0.9132 & 0.8517 & 0.8305 \\
\hline
Struct & 0.3239 & 0.4333 & 0.9543 & 0.9326 \\
\hline
Prefix & 0.9114 & 0.9025 & 0.8332 & 0.7676 \\
\hline
StructCtx. & 0.9429 & 0.9876 & - & - \\
\hline
Self-node & 0.3840 & 0.5382 & - & - \\
\hline
all & 0.1316 & 0.0597 & 0.4590 & 0.3725 \\
\hline
\end{tabular}
}
\end{subtable}


\caption{Attention corruption study on \textbf{decoder cross-attention} from each decoder step to encodings of each input section. 
}
\vspace{-0.3cm}
\label{tbl:exp5.4+6.2}
\end{table}


\stepcounter{takeaway}
\paragraph{Takeaway~\thetakeaway~(Q3).} The model exhibits \textit{duplicative robustness} in the joint representation of text and structure. \textbf{\textit{Both encoder and decoder demonstrate proficient capabilities in fusing text information into structure, highlighting both the internal robustness of the model but also possibilities for compression}}.

\subsubsection{Attention Weights Information}
To delve deeper into the functionalities of attention, we examine the potential correlation between attention weights and actual interpretable information. In \textbf{Q4} we hypothesize that the model internally performs subtasks akin to those designed by humans. One of them is \textit{schema linking}, which aims to determine the \textit{relevance} of a node, i.e., whether this node should appear in the output SQL. 
For that, we explore the correlations between the distribution patterns of a column within encoder self-attention and the relevance of the column.
In detail, within each sample, we gather the attention weights from each column\footnote{For simplicity, in this study we only use the first token of each column as its representative.} to different input sections, across different layers and attention heads. We then compare patterns between relevant and non-relevant columns to gauge any distinctive behavior.

The results are presented in Table~\ref{tbl:exp4.1-L23}. 
We indeed observe distinctions between relevant and non-relevant columns. For specific heads and input sections, attention is consistently high for only one type of columns and low for the other. For example, head 8 shows a notably high attention to text and low attention to structural context for relevant nodes. For heads 10 and 11, attention to prefix token \#4 is markedly high for non-relevant nodes.

We further confirm the correlation between encoder self-attention weights and node relevance by directly utilizing the attention patterns as features for node relevance classification. Specifically, we focused on the attention positions (layer, head, section) with clear discrepancy between relevant and non-relevant nodes, as mentioned above, and collect the attention weights on these positions to form ``input features'' for each node. A logistic regressor (LR) was trained to make the relevance prediction using these features. We compared with the implicit predictions made by the full model, where nodes in the generated SQL were predicted as relevant.
The results can be found in Table~\ref{tbl:exp4.1-probing}. The accuracy and F1 scores of the LR are on par with those of the full model, and significantly better than simple heuristics such as predicting nodes with ``exact text match'' as relevant. This reaffirms that the encoder self-attention is tightly associated with node relevance, and that the encoder has successfully internalized the schema linking subtask.

\stepcounter{takeaway}
\paragraph{Takeaway~\thetakeaway~(Q3, Q4).} 
Encoder self-attention weights  carry distinguishing information about node relevance. \textbf{
\textit{This implies the ability of the encoder to perform the schema linking subtask.}}

\begin{table}[t]
\centering
\begin{subtable}{\linewidth}
    \centering
    \resizebox{\linewidth}{!}{%
    \begin{tabular}{|p{4.5cm}|>{\centering\arraybackslash}p{2cm}|>{\centering\arraybackslash}p{2cm}|}
    \hline
    Corrupted part & Weights & Logits \\\hline
    Enc.SA-only & 0.9071 & 0.6879 \\\hline
    Dec.XA-only & 0.9063 & 0.9132 \\\hline
    Enc.SA + Dec.XA & 0.6414 & 0.2987 \\\hline
    \end{tabular}
    }
\end{subtable}

\caption{Joint corrupting encoder self-attention (structure to text) and decoder cross-attention (to text), to highlight the duplicative robustness phenomenon.
}
\label{tbl:exp5.5+6.3}
\end{table}

\begin{table}[t]
\centering

\begin{subtable}{\linewidth}
\centering
\resizebox{\linewidth}{!}{%
\begin{tabular}{|l|c|c|c|c|}
\hline
 & {Head 7} & {Head 8} & {Head 10} & {Head 11}\\\hline
prefix\#0 & \textcolor{red}{0.55 / 0.06} & 0.01 / 0.01 & 0.04 / 0.01 & 0.00 / 0.04 \\\hline
prefix\#4 & 0.04 / 0.05 & 0.00 / 0.00 & \textcolor{blue}{0.01 / 0.45} & \textcolor{blue}{0.01 / 0.42} \\\hline
prefix\#8 & 0.01 / 0.01 & \textcolor{blue}{0.01 / 0.24} & 0.07 / 0.02 & 0.12 / 0.03 \\\hline
text & 0.00 / 0.01 & \textcolor{red}{0.73 / 0.06} & 0.05 / 0.01 & 0.00 / 0.00 \\\hline
self & 0.25 / 0.57 & 0.01 / 0.01 & 0.03 / 0.00 & 0.04 / 0.05 \\\hline
context & 0.10 / 0.24 & \textcolor{blue}{0.17 / 0.61} & \textcolor{red}{0.33 / 0.09} & 0.22 / 0.23 \\\hline
others & 0.01 / 0.01 & 0.06 / 0.05 & 0.27 / 0.07 & \textcolor{red}{0.52 / 0.15} \\\hline
\end{tabular}

}
\end{subtable}



\caption{Attention weights from a column to each section, averaged for all relevant / non-relevant columns. Values in \textcolor{red}{red} are high for relevant nodes, and in \textcolor{blue}{blue} are high for non-relevant nodes.
Due to space constraints we only show results for Encoder layer 23, on a subset of heads and prefix tokens, on the dev set. More results are provided in Table~\ref{tbl:exp4.1-L23-appendix}.
\vspace{-0.5cm}
}
\label{tbl:exp4.1-L23}
\end{table}

\begin{table}[t]
\centering
\resizebox{\linewidth}{!}{%
\begin{tabular}{|l|c|c|c|}
\hline
Relevance & Attn. + LR & Full model & Heuristics \\
\hline
Accuracy & 0.9729 & 0.9842 & 0.9403 \\
\hline
Precision & 0.8669 & 0.9316 & 0.8616 \\
\hline
Recall & 0.8535 & 0.8994 & 0.4634 \\
\hline
F1 & 0.8602 & 0.9152 & 0.6026 \\
\hline
\end{tabular}
}
\caption{Column relevance prediction results. 
The P/R/F1 is computed for ``relevant'' columns due to their sparsity (<10\%) among all columns.
}
\vspace{-0.3cm}
\label{tbl:exp4.1-probing}
\end{table}

\begin{table}[t]
\begin{subtable}{1.0\linewidth}
\centering
\resizebox{\linewidth}{!}{%
\begin{tabular}{|c|c|c|c||c|c|}
\hline
Module & Corruption type & Layers & Section & Exact & Exec \\\hline
Enc.SA & Logits & Low & $S\rightarrow T$ & 0.6538 & 0.6654 \\\hline
Enc.SA & Logits & Low & $T\rightarrow S$ & 0.6518 & 0.6721 \\\hline
Enc.SA & Logits & Low & $T\leftrightarrow S$ & 0.6422 & 0.6634 \\\hline
Enc.SA & Logits & High & $S\rightarrow T$ & 0.4072 & 0.4362 \\\hline
Enc.SA & Logits & all & $S\rightarrow T$ & 0.2234 & 0.2369 \\\hline
Enc.SA & Weights & all & $S\rightarrow T$ & 0.5145 & 0.5387 \\\hline\hline
Dec.XA & Logits & all & Text & 0.0706 & 0.0812 \\\hline
Dec.XA & Logits & all & Struct & 0.0648 & 0.0638 \\\hline\hline
Dec.SA & Weights & all & all & 0.0000 & 0.0000 \\\hline\hline
- & (Clean) & - & - & 0.6692 & 0.6809 \\\hline
\end{tabular}
}
\end{subtable}

\caption{End-to-end SQL performance with different attention corruption settings. Rows are selected for discussion; full results are available in Table~\ref{tbl:exp-A1.*-appendix} in the appendix. SA: self-attention; XA: cross-attention.}
\vspace{-0.3cm}
\label{tbl:exp-A1.*}
\end{table}

\subsubsection{End-to-End SQL Performance and Error Analysis}
To verify the above findings, which are based on token-level prediction results, we extended our experiments using the same corruption settings to evaluate the \textit{end-to-end SQL prediction} performance. SQL predictions are measured by the Exact Match and Execution Match metrics, in line  with the original Spider leaderboard~\cite{yu2018spider}.
In these experiments, we introduced corruptions on different layer ranges and targeting specific sections such as the encoder self-attention between the text and structure, and decoder cross-attention to text and structure. Additionally, we added \textit{decoder self-attention} corruption, which yielding interesting findings.

The results are provided in Table~\ref{tbl:exp-A1.*}. 
For encoder self-attention (Enc.SA), the trends are consistent with previous observations. For example, the performance of ``logits'' corruption is much lower than ``weights'' ($0.2369$ vs. $0.5387$ on Exec-match).
Interestingly, corruption on the lower layers resulted in almost no decrease in performance, even for section ``$T\leftrightarrow S$'' ($0.6809\rightarrow 0.6634$), hinting at opportunities for model pruning.

\begin{figure}[t]
\centering
\includegraphics[width=0.9\linewidth]{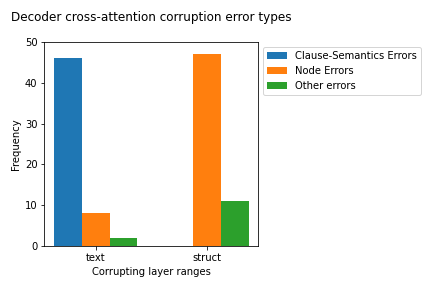}
\vspace{-0.3cm}
\caption{Error type analysis on \textbf{decoder cross-attention} corruption on the text or structure part. 
} 
\vspace{-0.1cm}
\label{fig:exp-A1.1.1-error-analysis-htypes}
\end{figure}

For decoder cross-attention (Dec.XA), introducing corruption to either text or structure drastically reduces performance. This is attributable to the compromised ability to predict syntax tokens and structure nodes, respectively. To further understand the actual behavior of the model \textbf{(Q4)}, we conducted  manual error analysis on a subset of $50$ samples.
The results are shown in Figure~\ref{fig:exp-A1.1.1-error-analysis-htypes}, with supplementary details provided in Table~\cref{tbl:exp-A1.1.1-error-analysis-criteria,tbl:exp-A1.1.1-error-analysis-break-down} in the appendix. We observe that when cross attention to text is blocked, the predominant errors are ``clause-semantic errors,'' most commonly missing a condition or aggregation function. On the other hand, blocking structure section primarily results in node selection errors, where the model hallucinates on node names. 
This finding verifies the specialized capabilities of the decoder for SQL syntax prediction and node selection, functioning independently of each other. 
This reinforces the conclusion from Takeaway \textbf{\thetakeaway} for \textbf{Q4} that the model mirrors human-designed pipelines, including schema linking, syntax prediction, and node selection.

\stepcounter{takeaway}
\paragraph{Takeaway~\thetakeaway~(Q4)} The model shows the \textbf{\textit{ability to perform different subtasks corresponding to human-designed pipelines: 
schema linking in the encoder self-attention, syntax prediction in decoder cross-attention to text, and node selection in decoder cross-attention to structure}}.

\begin{figure}[t]
\centering
\includegraphics[width=0.9\linewidth]{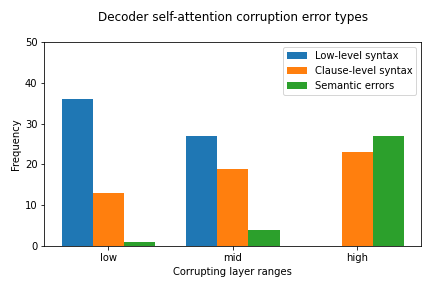}\vspace{-0.1cm}
\vspace{-0.3cm}
\caption{Error type analysis on \textbf{decoder self-attention} corruption on various layer ranges. 
} 
\vspace{-0.2cm}
\label{fig:exp-A1.2.0-error-analysis-htypes}
\end{figure}

\begin{table}[t]
\begin{subtable}{1.0\linewidth}
\centering
\resizebox{\linewidth}{!}{%
\begin{tabular}{|l|}
\hline
Q: \textbf{What is the average horsepower for all cars produced before 1980 ?} \\\hline
Pred: \texttt{SELECT avg(horsepower) FROM cars\_data WHERE year \textcolor{red}{\textbf{prior}} 1980} \\\hline
Gold: \texttt{SELECT avg(horsepower) FROM cars\_data WHERE year \textcolor{blue}{<} 1980} \\\hline

\addlinespace[4pt]\hline
Q: \textbf{What are the codes of template types that have fewer than 3 templates? } \\\hline
Pred: \texttt{SELECT template\_type\_code FROM templates GROUP BY} \\\addlinespace[-2pt]
\texttt{template\_type\_code HAVING COUNT(*) \textcolor{red}{\textbf{less than}} 3} \\\hline
Gold: \texttt{SELECT template\_type\_code FROM templates GROUP BY} \\\addlinespace[-2pt]
\texttt{template\_type\_code HAVING COUNT(*) \textcolor{blue}{<} 3} \\\hline

\addlinespace[4pt]\hline
Q: \textbf{What are the names of airports in Aberdeen?} \\\hline
Pred: \texttt{SELECT airportname FROM airports WHERE city \textcolor{red}{\textbf{is}} Aberdeen} \\\hline
Gold: \texttt{SELECT airportname FROM airports WHERE city \textcolor{blue}{=} "Aberdeen"} \\\hline

\addlinespace[4pt]\hline
Q: \textbf{List the title of all cartoons in alphabetical order.} \\\hline
Pred: \texttt{SELECT title FROM cartoon \textcolor{red}{\textbf{arranged alphabetically}}} \\\hline
Gold: \texttt{SELECT title FROM cartoon \textcolor{blue}{ORDER BY title}} \\\hline


\addlinespace[4pt]\hline
Q: \textbf{How many type of governments are in Africa?} \\\hline
Pred: \texttt{SELECT COUNT(\textcolor{red}{\textbf{different}} governmentform)} \\\addlinespace[-2pt]
\texttt{FROM country WHERE continent = Africa} \\\hline
Gold: \texttt{SELECT COUNT(\textcolor{blue}{DISTINCT} governmentform)} \\\addlinespace[-2pt]
\texttt{FROM country WHERE continent = "Africa"} \\\hline
\end{tabular}
}
\end{subtable}

\caption{Samples of corrupting \textbf{decoder self-attention} on \textbf{high layers}. In \textcolor{red}{\textbf{red}} are natural phrases generated by the model that semantically match the SQL syntax counterparts in \textcolor{blue}{blue}.}
\label{tbl:dec-self-high-samples}
\end{table}

\paragraph{Decoder Self-attention Study.}
In Takeaway~\textbf{\thetakeaway}, we confirmed the model's subtask handling capabilities and identified corresponding submodules. Delving deeper, we aim to pinpoint the layers where these processes occur, relating back to \textbf{Q2} concerning the storage of information within the layer dimension. Hypothetically, different layer ranges in the decoder have distinct responsibilities. Intuitively, lower layers would focus more on syntax prediction, while higher layers would concentrate more on node selection, reflecting the intrinsic order of these subtasks.

To verify this, we conducted additional experiments by corrupting the \textit{decoder self-attention}, effectively blocking all incoming information from previous timesteps in the decoder, within different layer ranges. The goal was to identify the information that is either already available or still missing at each layer range. The overall results are presented in Table~\ref{tbl:exp-A1.*}, where, as expected, the end-to-end performance significantly deteriorated. For deeper insights, we performed manual error analysis on each layer range, including low, middle, and high, on a subset of $50$ samples.
The results are shown in Figure~\ref{fig:exp-A1.2.0-error-analysis-htypes}, with more details in Table~\cref{tbl:exp-A1.2.0-error-analysis-criteria,tbl:exp-A1.2.0-error-analysis-break-down} in the appendix.
We observe a clear spectrum of error types distributed across the corrupted layer ranges. For low layers, the errors are predominantly ``low-level syntax errors'' such as unpaired brackets or quotes. For middle layers, many errors are ``clause-level errors'' where missing clauses or operators invalidate the SQL. For high layers, the SQL predictions are generally better formed syntactically, and the errors tend to pertain to higher-level semantics of the SQL. 
Interestingly, often the SQL error is confined to a clause where the SQL grammars are replaced by \textit{natural language phrases with similar semantics}. Examples of this phenomenon are provided in Table~\ref{tbl:dec-self-high-samples}. This intriguing observation is a strong indicator that the model has learned to align the semantics of SQL with natural language, and these representations are already obtained within the lower layers.
Meanwhile, the overall behavior discrepancies between layer ranges support our intuitive hypotheses on the distinct functionalities of different layer ranges in the decoder.

\stepcounter{takeaway}
\paragraph{Takeaway~\thetakeaway~(Q2, Q4)} The decoder follows human intuitions that low layers focus more on syntax prediction and high layers focus more on node selection. 
Remarkably, we found that \textbf{\textit{the model   learns to align the semantics of SQL with natural language, despite no training on naturalized SQL versions}}, such as SemQL or NatSQL~\cite{irnet:dblp:conf/acl/guozgxllz19,natsql:DBLP:conf/emnlp/GanCXPWDZ21}. This suggests that the model learns meaningful knowledge rather than merely exploiting spurious correlations in the dataset.

\section{Conclusion}

We conducted a comprehensive study on the internal behavior of an encoder-decoder  language model,  specifically T5,  text-to-SQL parser. Through both probing  and  manipulation of internal states, we provide insight into various aspects such as the information transfer between encoder and decoder, the storage of crucial data within the model, the functions of attention mechanisms, the process of modality fusion, the internal processing pipeline of the model, and the intrinsic alignment of SQL and natural language semantics.  Our findings can  inform and guide future research in text-to-SQL and related structured data representation tasks.

\section{Limitations}
Our study  is limited to one type of  pretrained language model architecture and did not consider  a broader spectrum of models. The scope of our study was also limited by the computational resources required for analyzing larger models and the availability of model intermediates. Thus, future research directions include:
1) Exploring similar  studies across different pretrained language model architectures, such as the widely adopted decoder-only models~\cite{DBLP:conf/nips/BrownMRSKDNSSAA20}. ii) Examining the impact of model scaling, both in terms of increased parameters and data volume, following the insights from scaling laws~\cite{DBLP:conf/nips/HoffmannBMBCRCH22}.
iii) Extending the study to other structured data tasks, such as speech-to-SQL ~\cite{DBLP:conf/icassp/ShaoKN23} or text-to-plots ~\cite{DBLP:conf/acl/ShaoN20,wang2021interactivepm}.
and a variety of structured data sources such as knowledge graphs~\cite{DBLP:conf/webdb/NakasholeTW10,DBLP:conf/wsdm/NakasholeTW11,DBLP:phd/dnb/Nakashole13,DBLP:conf/naacl/NakasholeW12,DBLP:journals/sigmod/NakasholeWS13,DBLP:conf/aaai/MitchellCHTBCMG15,DBLP:conf/icassp/0003RN17,opendialkg:Moon2019OpenDialKGEC,diffkg:Tuan2022TowardsLI} and tables~\cite{tabert:Yin2020TaBERTPF,tapas:Herzig2020TaPasWS,grappa:yu2021grappagp,tableformer:Yang2022TableFormerRT}.

Our study was  carried out only on the Spider dataset for text-to-SQL. However, we do not consider  this to be a  significant limitation, since our objective was to interpret the model behavior rather than proposing and validating a novel model.


\bibliography{mybib}
\bibliographystyle{acl_natbib}

\clearpage
\appendix
\section{Details of Dataset and Model}
\paragraph{Spider Dataset} The Spider dataset has 7000 samples in the training set and 1034 samples in the dev set~\cite{yu2018spider}. Unless mentioned otherwise, our analytical studies were conducted on the full dev set.

\paragraph{Text-to-SQL Model} The model we study is T5-large with prefix-tuning, implemented in the USKG project~\cite{uskg:xie2022unifiedskgua}. Regarding the model size, it is almost the same as T5-large which has around 770M parameters.

\paragraph{Data Processing} The tokenizer we use is the T5 tokenizer imported from HuggingFace, consistent with the model.

\paragraph{Probing Models} For logistic regression, we use the Logistic Regression class from the Scikit-learn package~\cite{scikit-learn}, with hyperparameter $C=1.0$. For neural models including MLP (for the LP probing task) and T5-decoder (for the NR probing task), we use AllenNLP with Pytorch backend~\cite{Gardner2017AllenNLP}. The MLP probe for the LP task has 2 linear layers with a middle dimensionality of 64 and activation function LeakyReLU (slope = $0.01$). It is trained with Adam with initial learning rate $1e^{-4}$. The "probe decoder" for the NR task uses T5-large decoder architecture and pretrained parameters (without prefix-tuning). It is fine-tuned with Adam with initial learning rate $1e^{-5}$.

\section{Extra Details of Analysis}
\subsection{Probing Study}
For Link Prediction (LP), we made adjustments to balance the frequencies of relation labels, given that there are multiple dominant classes with "default" labels (indicating no specific relation).
For each input sample, which consists of a textual sentence and its corresponding structural input, we use only \( K \) node-pairs from each relation class. {In our study, we simply set \( K = 1 \).}

\subsection{Causal Tracing Study}
When the ground truth unit is multi-token, no matter node name or syntax tokens, we compute the probability of each sub-token in the ground truth with teacher-forcing, and compute the minimum probability (bottleneck) among all sub-tokens as the probability of the entire unit.

When we study behavior of columns in the input, such as the effect of "self-node" or the attention to other sections, we exclude the column whose name appears more than once in the structured input, since it is non-trivial to define "self-node" in such cases. In detail, it is unclear whether another node with the exact same name should be considered as "self-node".

We corrupt representation vectors to zero-vectors instead of adding a random noise vector as done in~\cite{rome:meng2022locatingae}. This is because we observed in our preliminary studies that adding a random noise does not change the model prediction in most cases, different from the behaviors of the studied GPT-2 model in~\cite{rome:meng2022locatingae}.

\section{Extra Results}
\subsection{Attention Blocking Effect}
In comparing the outcomes of corrupting "weights" versus "logits" in Table~\cref{tbl:exp5.3+6.1,tbl:exp5.4+6.2}, an interesting trend emerges. 
In Table~\ref{tbl:exp5.3+6.1},when corrupting "$S\rightarrow T$" (and similarly "$T\leftrightarrow S$") in encoder self-attention, the "logits" corruption caused a higher detriment than "weights" ($0.9071$ vs. $0.6879$). This means that reallocating encoder self-attention weights to other less-attended sections can severely hamper performance.
Conversely, Table~\ref{tbl:exp5.4+6.2} does not display the same pattern. When corrupting decoder cross-attention to "Text", the accuracy under the "logits" setting is not inferior to "weights" but slightly superior ($0.9063$ vs. $0.9132$). For other sections like "Struct" or "Self-node", the "logits" accuracy is also higher than the "weights" setting. 
Besides, such patterns are only observed for column prediction and not for syntax tokens.

We attribute these trends to what we term the \textit{blocking effect} of encoder self-attention, where a high attention weights serves a dual purpose: transmitting information from that section while simultaneously inhibiting other sections from conveying information.
This effect is also verified by our finding on the correlation of attention weights and node relevance, which proves the attention weights carry distinguishing information. This explains the lowered performance when reallocating the attention to other sections, and justifies the existence of blocking effect in encoder self-attention.


\subsection{Attention per Layer Range}
\begin{figure*}[t]
\centering
    \begin{subfigure}[b]{0.5\linewidth}
    \includegraphics[width=\linewidth]{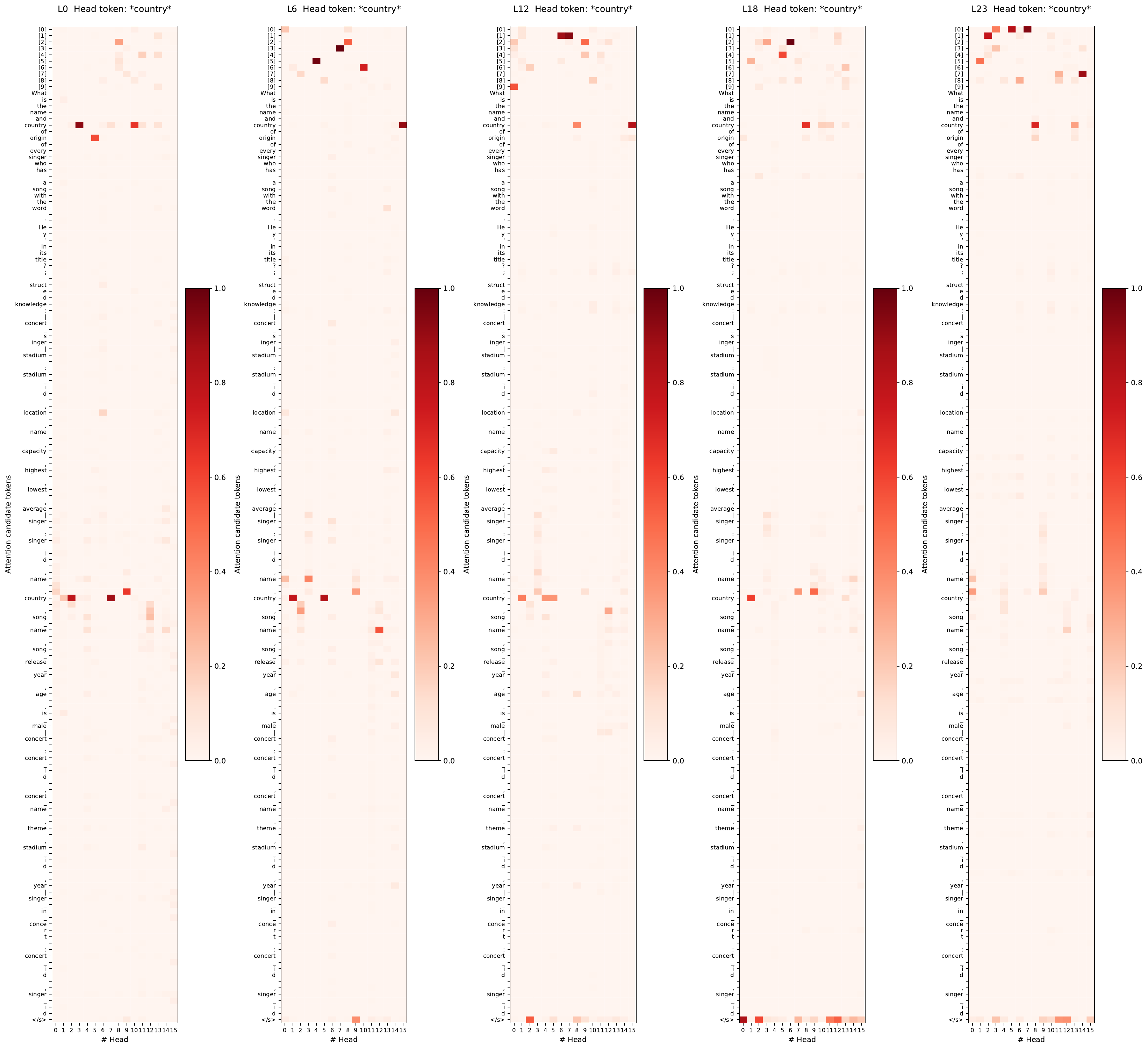}\vspace{-0.1cm}
    \caption*{(column-1)}\label{fig:exp4.0-1}
    \end{subfigure}

    \begin{subfigure}[b]{0.6\linewidth}
    \includegraphics[width=\linewidth]{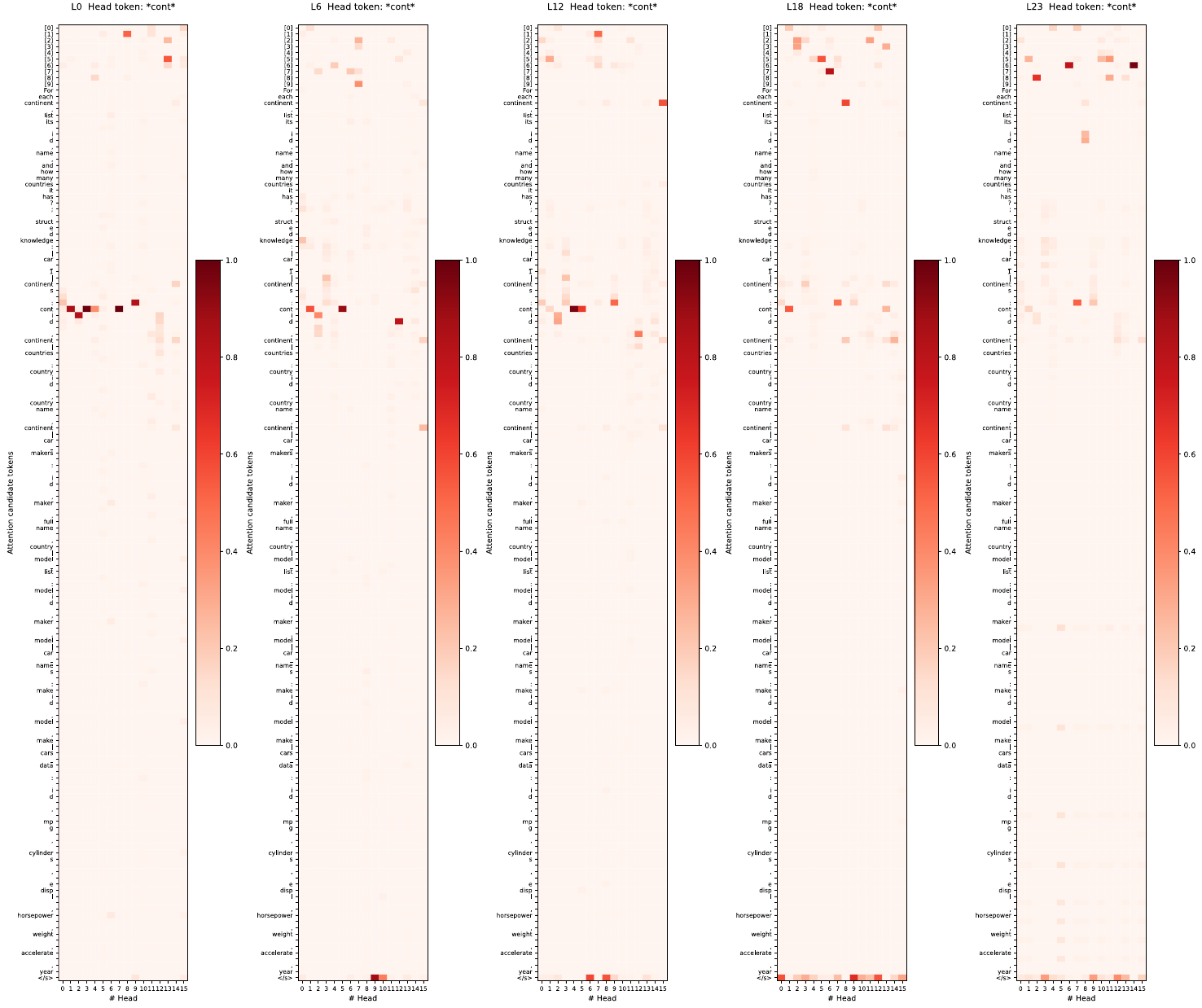}\vspace{-0.1cm}
    \caption*{(column-2)}\label{fig:exp4.0-2}
    \end{subfigure}

    \begin{subfigure}[b]{0.5\linewidth}
    \includegraphics[width=\linewidth]{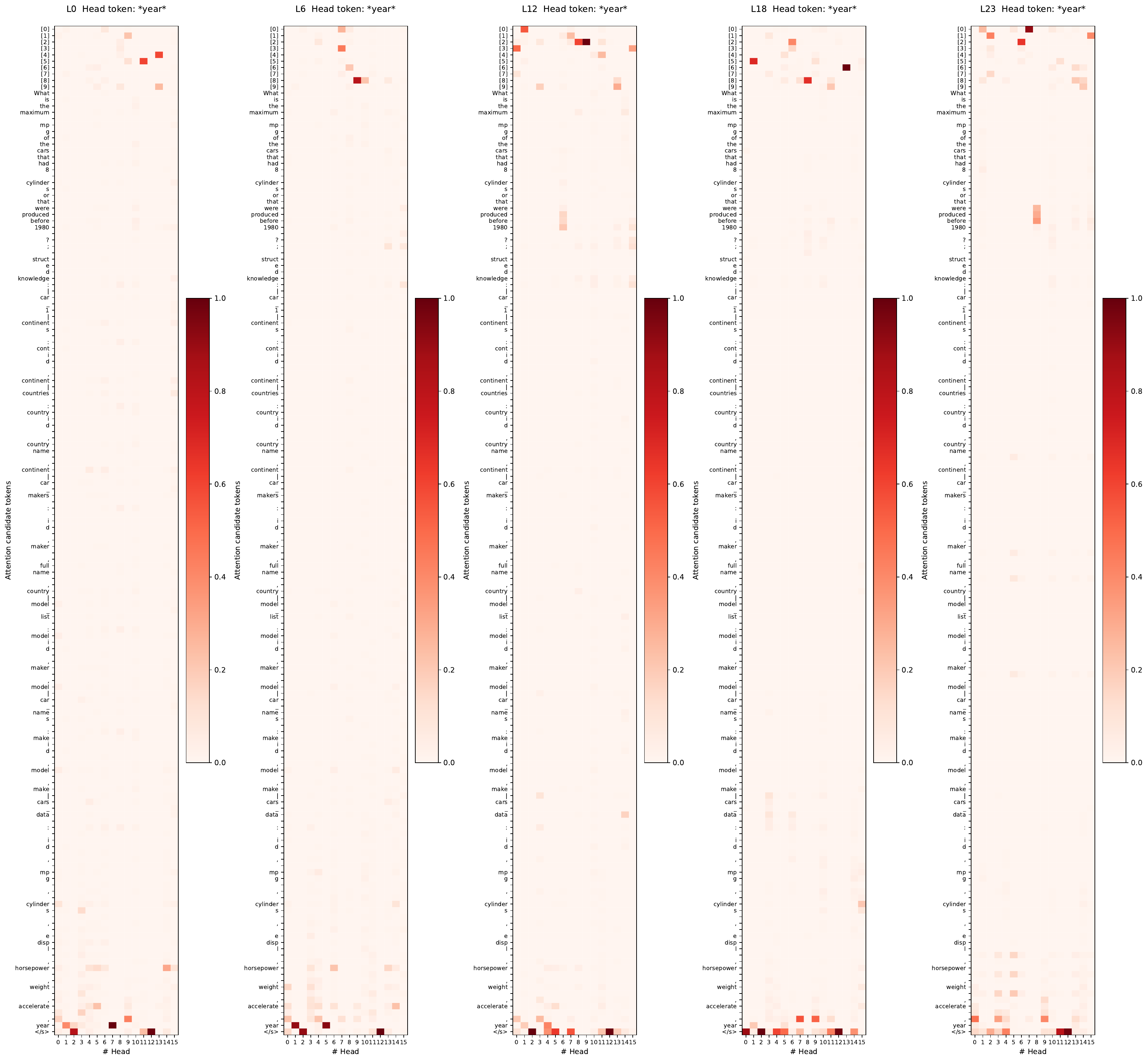}\vspace{-0.1cm}
    \caption*{(column-3)}\label{fig:exp4.0-3}
    \end{subfigure}
\caption{Encoder self-attention direct visualization.} 
\label{fig:exp4.0}
\end{figure*}

Exploring another aspect of \textbf{Q3} (attention functionalities), we check the importance of attention within different layer ranges. It also touches on \textbf{Q4} regarding the inner pipeline of the model.
Referring back to Table~\cref{tbl:exp5.3+6.1,tbl:exp5.4+6.2}, it is evident that for the encoder self-attention, corruption in low layers yields a substantially milder effect compared to high layers. This trend is consistent for both structure nodes and syntax prediction.
This suggests that, within the encoder, the high-layer self-attention takes an overall lead on the contextualization process. We further conducted direct observations on a subset of samples, examining the attention distribution from a column token to all tokens. We found that low-layer self-attention is more "self-concentrated", meaning it is largely attending to the self-node tokens. In contrast, attention in high layers is generally more distributed towards other tokens. Relevant visualizations can be found in Figure~\ref{fig:exp4.0}. 
Another evidence is from the attention distribution patterns (Table~\ref{tbl:exp4.1-L23-appendix}). High layers exhibit more distinctive attention positions than low layers. This further implies that high layers play a more pivotal role in gathering information.

For the decoder cross-attention, the comparison between low and high layers is less clear. One possible explanation is that different layer ranges have distinct responsibilities. We will further discuss this topic in later sections.

\section{Experiment Environment}
\paragraph{CPU:} 48 $\times$ Intel(R) Xeon(R) Gold 6136 CPU @ 3.00GHz
\paragraph{GPU:} 4 $\times$ NVIDIA GeForce GTX 1080 Ti
\paragraph{CUDA:} Version = 10.2
\paragraph{OS:} Ubuntu 16.04.6 LTS (Xenial)

\begin{table}[h]
\centering
\resizebox{\linewidth}{!}{%
\begin{tabular}{|l|c|c|c|}
\hline
Relation	& Freq	& T5-P-tuned	& T5-pretrained \\\hline
qq\_dist(-2)	& 500	& 0.7725	& 0.7367 \\\hline
qq\_dist(-1)	& 500	& 0.5164	& 0.5196 \\\hline
qq\_dist(0)	& 500	& 0.9159	& 0.9684 \\\hline
qq\_dist(1)	& 500	& 0.4990	& 0.5125 \\\hline
qq\_dist(2)	& 500	& 0.7787	& 0.7324 \\\hline
qc\_default	& 500	& 0.9098	& 0.8793 \\\hline
qt\_default	& 500	& 0.9715	& 0.9547 \\\hline
cq\_default	& 500	& 0.8877	& 0.8598 \\\hline
cc\_default	& 500	& 0.6264	& 0.6385 \\\hline
cc\_foreign\_key\_forward	& 471	& 0.8986	& 0.8925 \\\hline
cc\_foreign\_key\_backward	& 471	& 0.9104	& 0.9031 \\\hline
cc\_table\_match	& 500	& 0.6274	& 0.6141 \\\hline
cc\_dist(0)	& 500	& 0.9210	& 0.9440 \\\hline
ct\_default	& 500	& 0.7035	& 0.6549 \\\hline
ct\_primary\_key	& 465	& 0.9110	& 0.8665 \\\hline
ct\_table\_match	& 500	& 0.7522	& 0.7249 \\\hline
ct\_any\_table	& 500	& 0.9950	& 0.9980 \\\hline
tq\_default	& 500	& 0.9639	& 0.9660 \\\hline
tc\_default	& 500	& 0.6730	& 0.6046 \\\hline
tc\_primary\_key	& 465	& 0.8862	& 0.8580 \\\hline
tc\_table\_match	& 500	& 0.7543	& 0.6991 \\\hline
tc\_any\_table	& 500	& 0.9891	& 0.9960 \\\hline
tt\_default	& 471	& 0.5612	& 0.5571 \\\hline
tt\_foreign\_key\_forward	& 439	& 0.6683	& 0.6073 \\\hline
tt\_foreign\_key\_backward	& 439	& 0.6396	& 0.6369 \\\hline
tt\_dist(0)	& 500	& 0.9620	& 0.9320 \\\hline
qcCEM	& 264	& 0.8910	& 0.8673 \\\hline
cqCEM	& 264	& 0.8922	& 0.8729 \\\hline
qtTEM	& 199	& 0.9466	& 0.9300 \\\hline
tqTEM	& 199	& 0.9249	& 0.9193 \\\hline
qcCPM	& 281	& 0.8885	& 0.8405 \\\hline
cqCPM	& 281	& 0.9024	& 0.8364 \\\hline
qtTPM	& 51	& 0.8367	& 0.8200 \\\hline
tqTPM	& 51	& 0.8315	& 0.8444 \\\hline
qcNUMBER	& 81	& 0.9701	& 0.9405 \\\hline
cqNUMBER	& 81	& 0.9529	& 0.9277 \\\hline
qcTIME	& 15	& 0.8889	& 0.7500 \\\hline
cqTIME	& 15	& 0.8889	& 0.7692 \\\hline
qcCELLMATCH	& 125	& 0.8250	& 0.7410 \\\hline
cqCELLMATCH	& 125	& 0.7686	& 0.7903 \\\hline
\end{tabular}
}
\vspace{-0.1cm}
\caption{Link prediction (LP) probing results (F1-score) per relation type, using logistic regression (LR) as probing method. For detailed definitions of each relation, please refer to~\cite{ratsql:dblp:conf/acl/wangslpr20}.}
\vspace{-0.3cm}
\label{tbl:probing-LP-per-rel}
\end{table}

\begin{figure*}[t]
\centering
    \begin{subfigure}[b]{0.30\linewidth}
    \includegraphics[width=\linewidth]{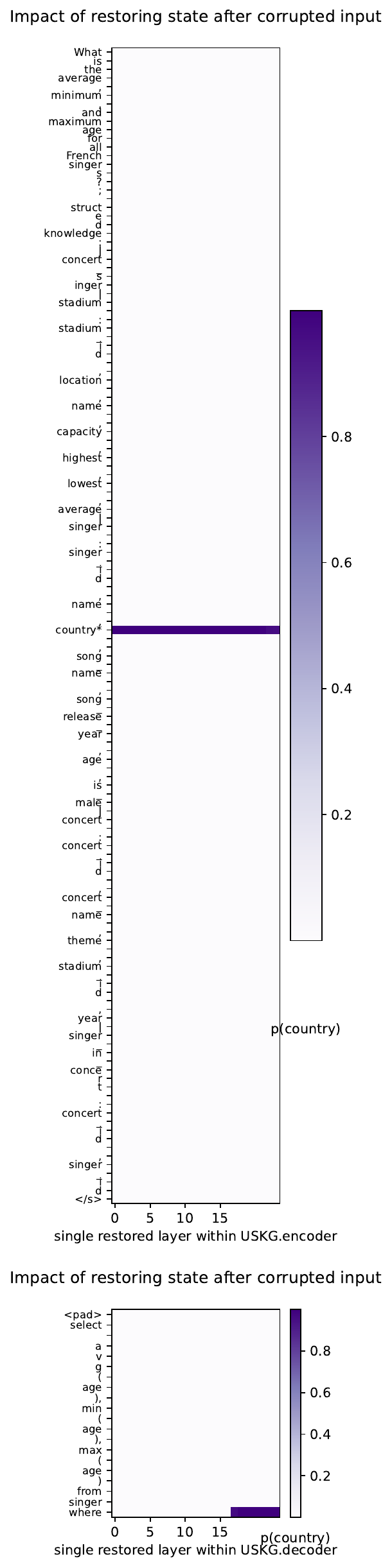}\vspace{-0.1cm}
    \caption*{(column-1)}\label{fig:exp1-a1-appendix}
    \end{subfigure}~
    \begin{subfigure}[b]{0.33\linewidth}
    \includegraphics[width=\linewidth]{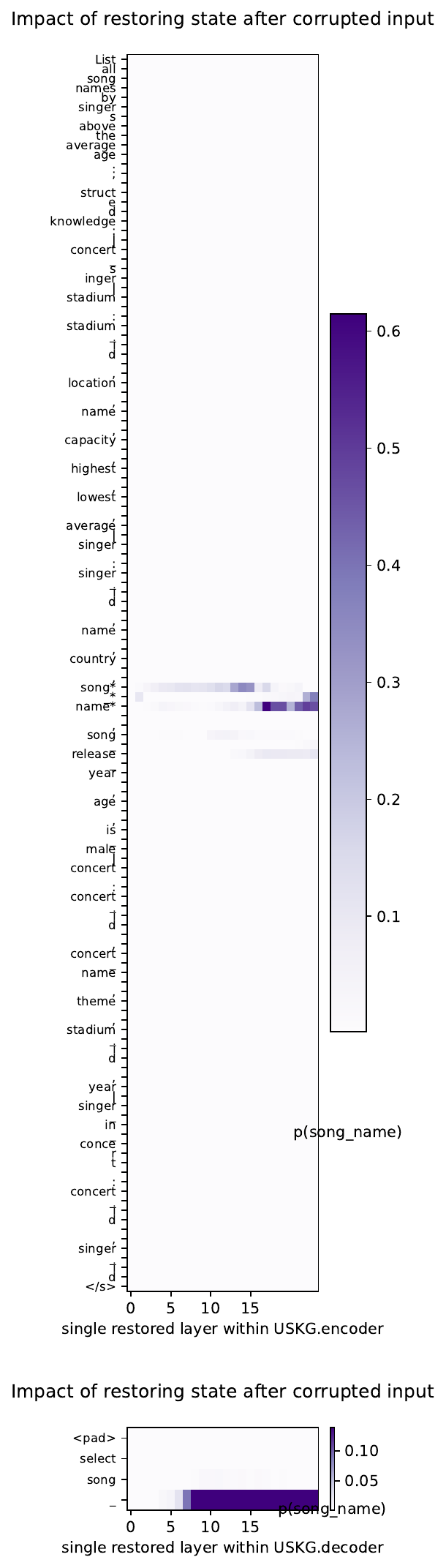}\vspace{-0.1cm}
    \caption*{(column-2)}\label{fig:exp1-a2-appendix}
    \end{subfigure}
\end{figure*}

\begin{figure*}[t]
\centering
    \begin{subfigure}[b]{0.39\linewidth}
    \includegraphics[width=\linewidth]{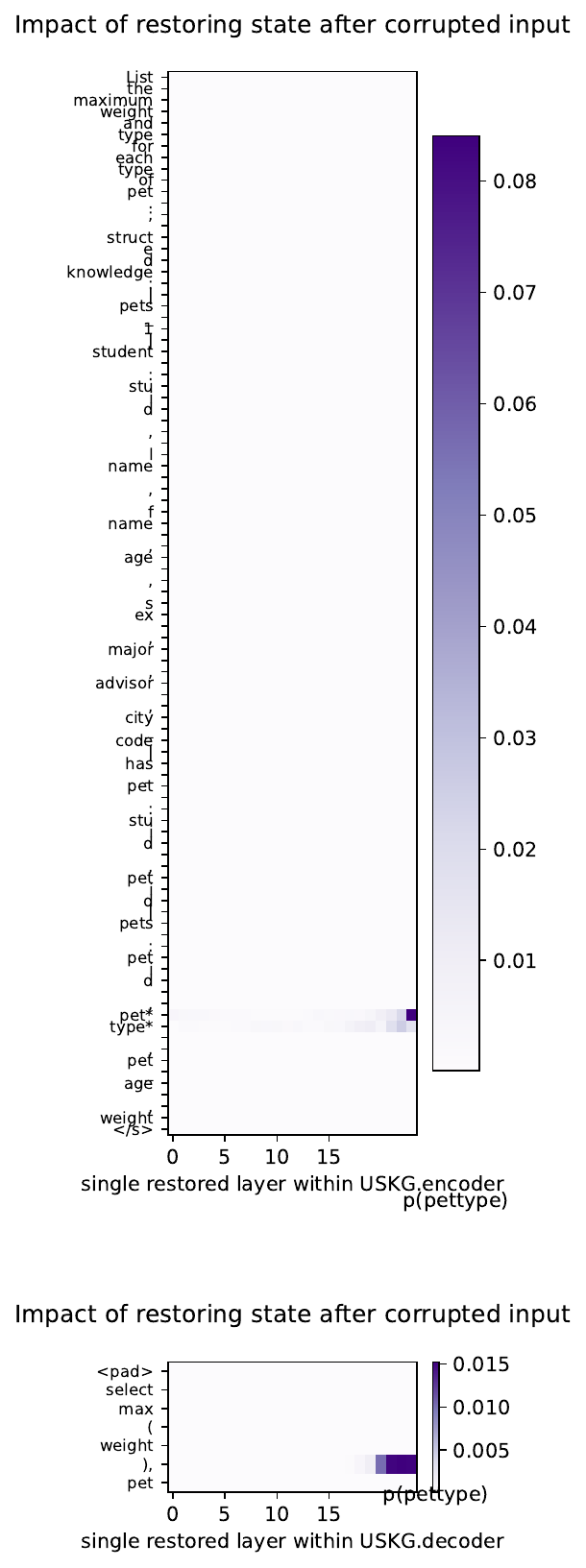}\vspace{-0.1cm}
    \caption*{(column-3)}\label{fig:exp1-b1-appendix}
    \end{subfigure}~
    \begin{subfigure}[b]{0.30\linewidth}
    \includegraphics[width=\linewidth]{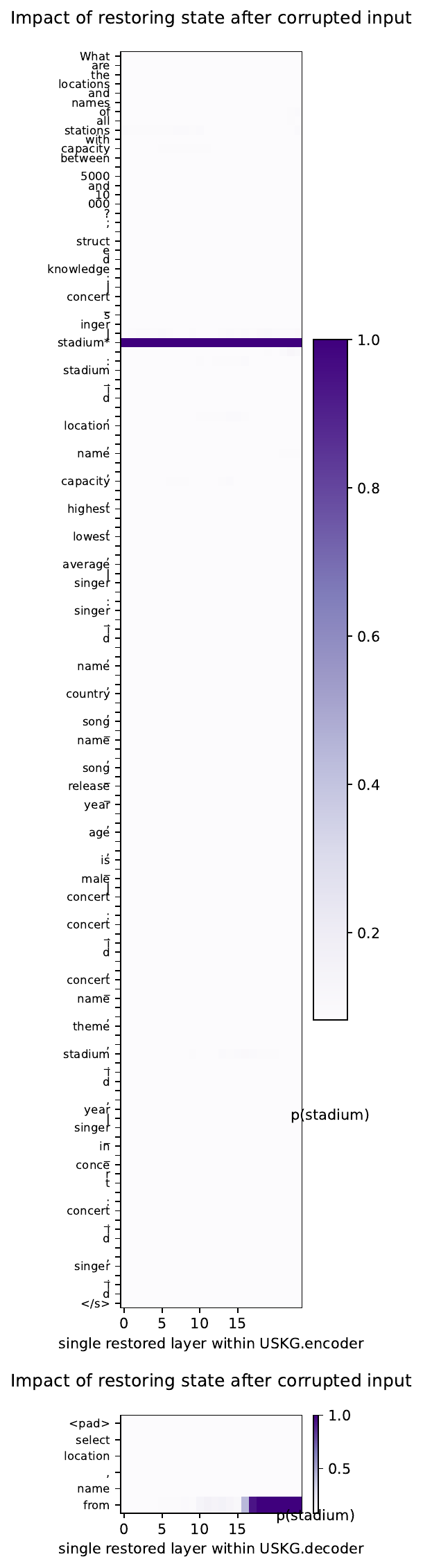}\vspace{-0.1cm}
    \caption*{(table-1)}\label{fig:exp1-b2-appendix}
    \end{subfigure}
\caption{Encoder state restoration effectiveness. Multi-token nodes are usually harder to recover by restoring a single state. Supplementary for Figure~\ref{fig:exp1}.}
\label{fig:exp1-appendix}
\end{figure*}

\begin{table}[t]
\centering
\begin{subtable}{1.0\linewidth}
\resizebox{\linewidth}{!}{%
    \centering
    \begin{tabular}{|l|c|c|c|}
    \hline
    Column  & Clean Text & DC. Text & Crpt. Text \\
    \hline
    Clean struct & 0.9905 & 0.9611 & 0.8505 \\
    \hline
    DC. struct & 0.6152 & 0.4952 & 0.0489 \\
    \hline
    Crpt. struct & 0.3916 & 0.3469 & 0.0241 \\
    \hline
    \end{tabular}
    }
    \vspace{0.3cm}
    \label{tbl:exp2.1.1-col-appendix}
\end{subtable}
\begin{subtable}{1.0\linewidth}
\resizebox{\linewidth}{!}{%
    \centering
    \begin{tabular}{|l|c|c|c|}
    \hline
    Table  & Clean Text & DC. Text & Crpt. Text \\
    \hline
    Clean Struct & 0.9906 & 0.9802 & 0.9687 \\
    \hline
    DC. Struct    & 0.4472 & 0.3965 & 0.1020 \\
    \hline
    Crpt. Struct & 0.2241 & 0.1958 & 0.0015 \\
    \hline
    \end{tabular}
    }
    \label{tbl:exp2.1.1-tab-appendix}
\end{subtable}
\vspace{-0.3cm}
\caption{Results of "dirty context encodings".
"DC." means "dirty context", i.e. this section is encoded with its own embeddings clean but the other part embeddings corrupted; the other part's final encodings are restored to the clean state. Effectively, in this setting we obtain encodings without information from the context. "Crpt." means "corrupted", i.e. the input embeddings of this section is corrupted.}
\label{tbl:exp2.1.1-appendix}
\end{table}

\begin{table}[t]
\centering
\resizebox{\linewidth}{!}{%
\begin{tabular}{|l|c|c|c|c|c|}
\hline
            & Both+ & Clean+ & SCC+ & None  & Total \\\hline
Column       & 1333 & 103    & 66   & 500   & 2002  \\\hline
Table        & 1424 & 123    & 105  & 31    & 1683  \\\hline
Table alias  & 1509 & 191    & 64   & 275   & 2039  \\\hline
\end{tabular}
}
\vspace{-0.1cm}
\caption{Struct context corruption (SCC) effect. Clean+ means clean prediction is correct, with SCC it is wrong. Vice versa, SCC+ means clean prediction is wrong, but SCC makes it correct. We see the number of cases in which such corruption has positive / negative effects are on the same level, indicating that the two factors (information removal \& distraction removal) both exist in structure context corruption.}
\vspace{-0.3cm}
\label{tbl:exp3.2}
\end{table}

\begin{table*}[t]
\centering

\begin{subtable}{\linewidth}
\centering
\resizebox{\linewidth}{!}{%
\begin{tabular}{|l|*{16}{c|}}
\hline
Dev-L1 & Head 0 & Head 1 & Head 2 & Head 3 & Head 4 & Head 5 & Head 6 & Head 7 & Head 8 & Head 9 & Head 10 & Head 11 & Head 12 & Head 13 & Head 14 & Head 15 \\\hline
prefix\#0	& {0.00 / 0.00} &	{0.02 / 0.03} &	{0.04 / 0.03} &	{0.01 / 0.01} &	{0.00 / 0.00} &	{0.09 / 0.11} &	{0.02 / 0.02} &	{0.00 / 0.00} &	{0.01 / 0.01} &	{0.01 / 0.01} &	{0.00 / 0.00} &	{0.01 / 0.01} &	{0.00 / 0.00} &	{0.00 / 0.00} &	{0.04 / 0.05} &	{0.01 / 0.02} \\\hline
prefix\#1	& {0.00 / 0.00} &	{0.01 / 0.01} &	{0.00 / 0.00} &	{0.00 / 0.00} &	{0.02 / 0.01} &	{0.01 / 0.02} &	{0.03 / 0.03} &	{0.00 / 0.00} &	{0.00 / 0.00} &	{0.00 / 0.00} &	{0.00 / 0.00} &	{0.00 / 0.00} &	{0.00 / 0.00} &	{0.01 / 0.00} &	{0.05 / 0.05} &	{0.00 / 0.00} \\\hline
prefix\#2	& {0.00 / 0.00} &	{0.01 / 0.01} &	{0.00 / 0.00} &	{0.04 / 0.05} &	{0.09 / 0.08} &	{0.11 / 0.06} &	{0.01 / 0.01} &	{0.00 / 0.00} &	{0.01 / 0.01} &	{0.02 / 0.01} &	{0.00 / 0.00} &	{0.03 / 0.02} &	{0.01 / 0.01} &	{0.01 / 0.01} &	{0.01 / 0.02} &	{0.05 / 0.05} \\\hline
prefix\#3	& {0.01 / 0.00} &	{0.00 / 0.00} &	{0.00 / 0.00} &	{0.00 / 0.00} &	{0.03 / 0.03} &	{0.03 / 0.02} &	{0.04 / 0.03} &	{0.00 / 0.00} &	{0.03 / 0.03} &	{0.01 / 0.01} &	{0.00 / 0.00} &	{0.00 / 0.00} &	{0.00 / 0.00} &	{0.00 / 0.00} &	{0.01 / 0.01} &	{0.02 / 0.04} \\\hline
prefix\#4	& {0.01 / 0.01} &	{0.01 / 0.01} &	{0.01 / 0.01} &	{0.00 / 0.00} &	{0.05 / 0.04} &	{0.02 / 0.02} &	{0.02 / 0.02} &	{0.00 / 0.00} &	{0.01 / 0.02} &	{0.00 / 0.00} &	{0.00 / 0.00} &	{0.00 / 0.01} &	{0.00 / 0.00} &	{0.00 / 0.00} &	{0.02 / 0.02} &	{0.02 / 0.03} \\\hline
prefix\#5	& {0.02 / 0.02} &	{0.08 / 0.09} &	{0.00 / 0.00} &	{0.01 / 0.01} &	{0.04 / 0.04} &	{0.02 / 0.02} &	{0.03 / 0.03} &	{0.00 / 0.00} &	{0.01 / 0.00} &	{0.06 / 0.04} &	{0.00 / 0.01} &	{0.01 / 0.01} &	{0.00 / 0.00} &	{0.00 / 0.00} &	{0.01 / 0.01} &	{0.00 / 0.01} \\\hline
prefix\#6	& {0.00 / 0.00} &	{0.01 / 0.01} &	{0.00 / 0.00} &	{0.00 / 0.00} &	{0.00 / 0.00} &	{0.02 / 0.03} &	{0.00 / 0.00} &	{0.00 / 0.01} &	{0.03 / 0.03} &	{0.00 / 0.00} &	{0.00 / 0.00} &	{0.00 / 0.00} &	{0.00 / 0.00} &	{0.01 / 0.01} &	{0.04 / 0.03} &	{0.00 / 0.00} \\\hline
prefix\#7	& {0.00 / 0.00} &	{0.00 / 0.00} &	{0.00 / 0.00} &	{0.00 / 0.00} &	{0.03 / 0.04} &	{0.06 / 0.02} &	{0.02 / 0.01} &	{0.00 / 0.00} &	{0.08 / 0.09} &	{0.00 / 0.00} &	{0.00 / 0.00} &	{0.07 / 0.08} &	{0.00 / 0.00} &	{0.00 / 0.00} &	{0.02 / 0.01} &	{0.01 / 0.01} \\\hline
prefix\#8	& {0.05 / 0.06} &	{0.06 / 0.05} &	{0.01 / 0.02} &	{0.00 / 0.00} &	{0.00 / 0.00} &	{0.02 / 0.03} &	{0.00 / 0.00} &	{0.00 / 0.00} &	{0.00 / 0.00} &	{0.00 / 0.00} &	{0.00 / 0.00} &	{0.01 / 0.02} &	{0.00 / 0.00} &	{0.00 / 0.00} &	{0.02 / 0.02} &	{0.00 / 0.01} \\\hline
prefix\#9	& {0.01 / 0.01} &	{0.01 / 0.01} &	{0.00 / 0.00} &	{0.00 / 0.00} &	{0.00 / 0.00} &	{0.01 / 0.01} &	{0.01 / 0.02} &	{0.00 / 0.00} &	{0.01 / 0.01} &	{0.00 / 0.00} &	{0.00 / 0.00} &	{0.01 / 0.02} &	{0.00 / 0.00} &	{0.00 / 0.00} &	{0.02 / 0.03} &	{0.02 / 0.05} \\\hline
text	& {0.05 / 0.01} &	{0.04 / 0.03} &	{0.00 / 0.00} &	{0.02 / 0.01} &	{0.03 / 0.01} &	{0.06 / 0.05} &	{0.12 / 0.09} &	{0.01 / 0.00} &	{0.21 / 0.11} &	{0.00 / 0.00} &	\textcolor{red}{0.41 / 0.02} &	{0.09 / 0.01} &	{0.00 / 0.00} &	\textcolor{red}{0.42 / 0.04} &	{0.05 / 0.03} &	\textcolor{red}{0.35 / 0.06} \\\hline
self	& {0.33 / 0.34} &	{0.44 / 0.39} &	{0.81 / 0.80} &	{0.20 / 0.22} &	{0.19 / 0.18} &	{0.09 / 0.08} &	{0.02 / 0.02} &	{0.26 / 0.22} &	{0.00 / 0.00} &	{0.63 / 0.66} &	{0.28 / 0.50} &	{0.11 / 0.08} &	{0.45 / 0.42} &	{0.13 / 0.33} &	{0.09 / 0.09} &	{0.01 / 0.01} \\\hline
context	& {0.50 / 0.51} &	{0.28 / 0.33} &	{0.09 / 0.10} &	{0.69 / 0.67} &	{0.50 / 0.53} &	{0.40 / 0.48} &	{0.50 / 0.57} &	{0.04 / 0.04} &	{0.40 / 0.49} &	{0.26 / 0.26} &	{0.30 / 0.47} &	{0.54 / 0.63} &	{0.39 / 0.45} &	{0.37 / 0.55} &	{0.49 / 0.53} &	{0.38 / 0.60} \\\hline
others	& {0.02 / 0.01} &	{0.02 / 0.02} &	{0.03 / 0.03} &	{0.03 / 0.01} &	{0.02 / 0.01} &	{0.06 / 0.06} &	{0.17 / 0.15} &	{0.69 / 0.73} &	{0.19 / 0.19} &	{0.00 / 0.00} &	{0.01 / 0.00} &	{0.10 / 0.10} &	{0.15 / 0.12} &	{0.04 / 0.05} &	{0.12 / 0.10} &	{0.12 / 0.11} \\\hline
\end{tabular}
}
\caption{Dev-L1}
\end{subtable}

\vspace{0.2cm}

\begin{subtable}{\linewidth}
\centering
\resizebox{\linewidth}{!}{%
\begin{tabular}{|l|*{16}{c|}}
\hline
Dev-L12 & Head 0 & Head 1 & Head 2 & Head 3 & Head 4 & Head 5 & Head 6 & Head 7 & Head 8 & Head 9 & Head 10 & Head 11 & Head 12 & Head 13 & Head 14 & Head 15 \\\hline
prefix\#0	& {0.00 / 0.00} &	{0.06 / 0.07} &	{0.04 / 0.06} &	{0.00 / 0.00} &	{0.00 / 0.00} &	{0.00 / 0.00} &	{0.00 / 0.00} &	{0.07 / 0.10} &	{0.00 / 0.00} &	{0.00 / 0.00} &	{0.01 / 0.06} &	{0.00 / 0.00} &	{0.00 / 0.00} &	{0.00 / 0.00} &	{0.00 / 0.00} &	{0.00 / 0.00} \\\hline
prefix\#1	& {0.00 / 0.01} &	{0.01 / 0.05} &	{0.00 / 0.00} &	{0.03 / 0.04} &	{0.00 / 0.00} &	{0.00 / 0.00} &	{0.10 / 0.05} &	{0.26 / 0.14} &	{0.00 / 0.01} &	{0.00 / 0.00} &	{0.01 / 0.02} &	{0.02 / 0.01} &	{0.00 / 0.01} &	{0.08 / 0.11} &	{0.00 / 0.00} &	{0.00 / 0.00} \\\hline
prefix\#2	& {0.16 / 0.12} &	{0.01 / 0.02} &	{0.01 / 0.01} &	{0.02 / 0.02} &	{0.00 / 0.00} &	{0.00 / 0.00} &	{0.10 / 0.05} &	{0.18 / 0.21} &	{0.01 / 0.01} &	{0.15 / 0.18} &	{0.00 / 0.00} &	{0.04 / 0.04} &	{0.05 / 0.10} &	{0.00 / 0.00} &	{0.00 / 0.00} &	{0.00 / 0.00} \\\hline
prefix\#3	& {0.08 / 0.03} &	{0.00 / 0.01} &	{0.00 / 0.01} &	{0.00 / 0.00} &	{0.00 / 0.00} &	{0.01 / 0.00} &	{0.04 / 0.02} &	{0.00 / 0.00} &	{0.01 / 0.01} &	{0.07 / 0.04} &	{0.03 / 0.01} &	{0.03 / 0.01} &	{0.00 / 0.01} &	{0.01 / 0.01} &	{0.01 / 0.02} &	{0.01 / 0.05} \\\hline
prefix\#4	& {0.07 / 0.04} &	{0.06 / 0.08} &	{0.01 / 0.01} &	{0.00 / 0.00} &	{0.00 / 0.00} &	{0.00 / 0.00} &	{0.05 / 0.12} &	{0.00 / 0.00} &	{0.00 / 0.00} &	{0.02 / 0.01} &	{0.02 / 0.01} &	{0.14 / 0.08} &	{0.00 / 0.00} &	{0.03 / 0.02} &	{0.01 / 0.01} &	{0.00 / 0.00} \\\hline
prefix\#5	& {0.05 / 0.05} &	{0.06 / 0.05} &	{0.00 / 0.00} &	{0.00 / 0.00} &	{0.00 / 0.00} &	{0.00 / 0.00} &	{0.07 / 0.03} &	{0.05 / 0.06} &	{0.00 / 0.00} &	{0.00 / 0.01} &	{0.00 / 0.01} &	{0.11 / 0.06} &	{0.00 / 0.00} &	{0.00 / 0.00} &	{0.01 / 0.01} &	{0.00 / 0.00} \\\hline
prefix\#6	& {0.01 / 0.01} &	{0.02 / 0.06} &	{0.13 / 0.09} &	{0.00 / 0.00} &	{0.00 / 0.00} &	{0.00 / 0.00} &	{0.00 / 0.00} &	{0.01 / 0.02} &	{0.00 / 0.00} &	{0.00 / 0.01} &	{0.01 / 0.02} &	{0.01 / 0.02} &	{0.01 / 0.01} &	{0.00 / 0.00} &	{0.03 / 0.02} &	{0.00 / 0.00} \\\hline
prefix\#7	& {0.02 / 0.01} &	{0.00 / 0.00} &	{0.01 / 0.01} &	{0.00 / 0.00} &	{0.00 / 0.00} &	{0.00 / 0.00} &	{0.01 / 0.00} &	{0.02 / 0.02} &	{0.00 / 0.00} &	{0.20 / 0.19} &	{0.01 / 0.01} &	{0.01 / 0.01} &	{0.00 / 0.00} &	{0.00 / 0.00} &	{0.01 / 0.01} &	{0.00 / 0.00} \\\hline
prefix\#8	& {0.00 / 0.00} &	{0.00 / 0.00} &	{0.00 / 0.00} &	{0.00 / 0.00} &	{0.00 / 0.00} &	{0.00 / 0.00} &	{0.00 / 0.00} &	{0.00 / 0.00} &	{0.00 / 0.00} &	{0.00 / 0.00} &	{0.04 / 0.02} &	{0.00 / 0.01} &	{0.00 / 0.00} &	{0.07 / 0.09} &	{0.00 / 0.00} &	{0.00 / 0.01} \\\hline
prefix\#9	& {0.11 / 0.12} &	{0.01 / 0.01} &	{0.00 / 0.00} &	{0.02 / 0.03} &	{0.00 / 0.00} &	{0.00 / 0.00} &	{0.00 / 0.01} &	{0.00 / 0.00} &	{0.00 / 0.00} &	{0.00 / 0.00} &	{0.00 / 0.00} &	{0.03 / 0.05} &	{0.02 / 0.01} &	{0.03 / 0.03} &	{0.00 / 0.00} &	{0.00 / 0.00} \\\hline
text	& {0.03 / 0.02} &	{0.02 / 0.02} &	{0.00 / 0.00} &	{0.00 / 0.00} &	{0.02 / 0.01} &	{0.04 / 0.02} &	{0.11 / 0.08} &	{0.01 / 0.01} &	\textcolor{red}{0.25 / 0.05} &	{0.00 / 0.00} &	{0.18 / 0.11} &	{0.01 / 0.01} &	{0.00 / 0.00} &	{0.14 / 0.08} &	{0.13 / 0.07} &	\textcolor{red}{0.71 / 0.18} \\\hline
self	& {0.18 / 0.20} &	{0.61 / 0.44} &	{0.33 / 0.39} &	{0.27 / 0.32} &	{0.57 / 0.49} &	{0.25 / 0.28} &	{0.01 / 0.01} &	{0.11 / 0.09} &	{0.00 / 0.00} &	{0.28 / 0.23} &	{0.02 / 0.01} &	{0.07 / 0.07} &	{0.44 / 0.38} &	{0.04 / 0.03} &	{0.15 / 0.13} &	{0.01 / 0.01} \\\hline
context	& {0.16 / 0.25} &	{0.10 / 0.14} &	{0.04 / 0.05} &	{0.63 / 0.58} &	{0.36 / 0.44} &	{0.40 / 0.44} &	{0.27 / 0.39} &	{0.06 / 0.07} &	{0.41 / 0.56} &	{0.07 / 0.06} &	{0.51 / 0.46} &	{0.45 / 0.56} &	{0.23 / 0.23} &	{0.46 / 0.45} &	{0.55 / 0.61} &	{0.18 / 0.44} \\\hline
others	& {0.10 / 0.13} &	{0.03 / 0.05} &	{0.41 / 0.36} &	{0.01 / 0.02} &	{0.04 / 0.06} &	{0.29 / 0.25} &	{0.22 / 0.23} &	{0.21 / 0.29} &	{0.31 / 0.36} &	{0.18 / 0.26} &	{0.15 / 0.24} &	{0.07 / 0.07} &	{0.24 / 0.24} &	{0.13 / 0.18} &	{0.10 / 0.11} &	{0.09 / 0.31} \\\hline
\end{tabular}
}
\caption{Dev-L12}
\end{subtable}

\vspace{0.2cm}

\begin{subtable}{\linewidth}
\centering
\resizebox{\linewidth}{!}{%
\begin{tabular}{|l|*{16}{c|}}
\hline
Dev-L23 & Head 0 & Head 1 & Head 2 & Head 3 & Head 4 & Head 5 & Head 6 & Head 7 & Head 8 & Head 9 & Head 10 & Head 11 & Head 12 & Head 13 & Head 14 & Head 15 \\\hline
prefix\#0 & {0.00 / 0.00} & {0.03 / 0.01} & {0.00 / 0.00} & {0.09 / 0.04} & {0.04 / 0.02} & {0.09 / 0.14} & {0.11 / 0.07} & \textcolor{red}{0.55 / 0.06} & {0.01 / 0.01} & {0.07 / 0.03} & {0.04 / 0.01} & {0.00 / 0.04} & \textcolor{blue}{0.06 / 0.32} & {0.02 / 0.00} & {0.00 / 0.00} & {0.01 / 0.00} \\\hline
prefix\#1 & {0.03 / 0.03} & {0.02 / 0.04} & {0.13 / 0.09} & {0.00 / 0.00} & {0.01 / 0.00} & {0.00 / 0.00} & {0.06 / 0.03} & {0.00 / 0.00} & {0.00 / 0.01} & {0.01 / 0.00} & {0.01 / 0.01} & {0.00 / 0.00} & {0.00 / 0.00} & {0.00 / 0.00} & {0.01 / 0.00} & {0.04 / 0.02} \\\hline
prefix\#2 & {0.08 / 0.01} & \textcolor{blue}{0.00 / 0.20} & {0.00 / 0.00} & {0.04 / 0.00} & {0.00 / 0.00} & {0.00 / 0.01} & {0.12 / 0.01} & {0.00 / 0.00} & {0.00 / 0.01} & {0.03 / 0.03} & {0.00 / 0.00} & {0.00 / 0.00} & {0.07 / 0.08} & {0.03 / 0.00} & {0.00 / 0.00} & {0.01 / 0.00} \\\hline
prefix\#3 & {0.00 / 0.00} & {0.00 / 0.03} & {0.14 / 0.01} & {0.10 / 0.00} & {0.01 / 0.01} & {0.00 / 0.00} & {0.00 / 0.00} & {0.00 / 0.00} & {0.00 / 0.00} & {0.00 / 0.04} & {0.05 / 0.01} & {0.03 / 0.01} & {0.00 / 0.00} & {0.00 / 0.00} & {0.12 / 0.09} & {0.00 / 0.00} \\\hline
prefix\#4 & {0.02 / 0.01} & {0.01 / 0.01} & {0.09 / 0.16} & {0.10 / 0.02} & {0.00 / 0.00} & {0.00 / 0.00} & {0.00 / 0.00} & {0.04 / 0.05} & {0.00 / 0.00} & {0.01 / 0.00} & \textcolor{blue}{0.01 / 0.45} & \textcolor{blue}{0.01 / 0.42} & {0.00 / 0.00} & {0.00 / 0.00} & {0.00 / 0.00} & {0.05 / 0.01} \\\hline
prefix\#5 & {0.00 / 0.00} & {0.17 / 0.04} & {0.01 / 0.00} & {0.01 / 0.00} & {0.00 / 0.00} & {0.00 / 0.00} & {0.04 / 0.02} & {0.01 / 0.04} & {0.00 / 0.00} & \textcolor{blue}{0.00 / 0.21} & {0.08 / 0.24} & {0.02 / 0.05} & {0.00 / 0.03} & {0.02 / 0.01} & {0.01 / 0.00} & {0.00 / 0.01} \\\hline
prefix\#6 & {0.01 / 0.00} & \textcolor{blue}{0.03 / 0.20} & {0.01 / 0.03} & {0.08 / 0.01} & {0.00 / 0.00} & {0.00 / 0.06} & {0.06 / 0.23} & {0.00 / 0.00} & {0.00 / 0.00} & {0.00 / 0.01} & {0.01 / 0.00} & {0.00 / 0.00} & {0.03 / 0.02} & {0.01 / 0.09} & {0.10 / 0.17} & {0.00 / 0.00} \\\hline
prefix\#7 & {0.01 / 0.01} & {0.01 / 0.11} & {0.02 / 0.00} & {0.00 / 0.00} & {0.00 / 0.03} & {0.00 / 0.00} & {0.00 / 0.00} & {0.00 / 0.00} & {0.00 / 0.00} & \textcolor{blue}{0.00 / 0.14} & {0.02 / 0.01} & {0.03 / 0.00} & {0.01 / 0.00} & {0.00 / 0.00} & \textcolor{red}{0.23 / 0.01} & {0.00 / 0.00} \\\hline
prefix\#8 & {0.02 / 0.05} & {0.01 / 0.01} & {0.06 / 0.14} & {0.00 / 0.00} & {0.04 / 0.02} & {0.00 / 0.00} & {0.03 / 0.01} & {0.01 / 0.01} & \textcolor{blue}{0.01 / 0.24} & {0.00 / 0.00} & {0.07 / 0.02} & {0.12 / 0.03} & {0.00 / 0.00} & \textcolor{red}{0.32 / 0.06} & {0.07 / 0.02} & {0.01 / 0.02} \\\hline
prefix\#9 & {0.00 / 0.00} & {0.00 / 0.00} & {0.00 / 0.00} & {0.00 / 0.00} & {0.00 / 0.00} & {0.00 / 0.00} & {0.01 / 0.00} & {0.00 / 0.00} & {0.00 / 0.00} & {0.00 / 0.00} & {0.01 / 0.07} & {0.00 / 0.00} & {0.00 / 0.00} & {0.00 / 0.00} & {0.07 / 0.00} & {0.00 / 0.00} \\\hline
text & {0.01 / 0.00} & {0.07 / 0.04} & {0.00 / 0.00} & {0.01 / 0.00} & {0.02 / 0.01} & {0.02 / 0.01} & {0.05 / 0.02} & {0.00 / 0.01} & \textcolor{red}{0.73 / 0.06} & {0.00 / 0.00} & {0.05 / 0.01} & {0.00 / 0.00} & {0.00 / 0.00} & {0.10 / 0.01} & {0.01 / 0.00} & {0.10 / 0.05} \\\hline
self & {0.46 / 0.40} & {0.09 / 0.04} & {0.41 / 0.37} & {0.07 / 0.21} & {0.29 / 0.24} & {0.03 / 0.11} & {0.02 / 0.03} & {0.25 / 0.57} & {0.01 / 0.01} & {0.25 / 0.26} & {0.03 / 0.00} & {0.04 / 0.05} & {0.28 / 0.19} & {0.14 / 0.11} & {0.10 / 0.10} & {0.04 / 0.02} \\\hline
context & {0.28 / 0.44} & {0.37 / 0.22} & {0.06 / 0.07} & {0.21 / 0.53} & {0.31 / 0.44} & {0.77 / 0.60} & {0.36 / 0.50} & {0.10 / 0.24} & \textcolor{blue}{0.17 / 0.61} & {0.29 / 0.13} & \textcolor{red}{0.33 / 0.09} & {0.22 / 0.23} & {0.15 / 0.12} & {0.25 / 0.59} & {0.26 / 0.56} & {0.46 / 0.68} \\\hline
others & {0.08 / 0.04} & {0.18 / 0.05} & {0.08 / 0.12} & {0.30 / 0.18} & {0.26 / 0.21} & {0.07 / 0.05} & {0.14 / 0.08} & {0.01 / 0.01} & {0.06 / 0.05} & {0.32 / 0.15} & {0.27 / 0.07} & \textcolor{red}{0.52 / 0.15} & {0.39 / 0.22} & {0.11 / 0.13} & {0.03 / 0.02} & {0.25 / 0.18} \\\hline
\end{tabular}
}
\caption{Dev-L23}
\end{subtable}

\vspace{0.2cm}

\begin{subtable}{\linewidth}
\centering
\resizebox{\linewidth}{!}{%
\begin{tabular}{|l|*{16}{c|}}
\hline
Train-L23 & Head 0 & Head 1 & Head 2 & Head 3 & Head 4 & Head 5 & Head 6 & Head 7 & Head 8 & Head 9 & Head 10 & Head 11 & Head 12 & Head 13 & Head 14 & Head 15 \\
\hline
prefix\#0 & {0.00 / 0.00} & {0.02 / 0.01} & {0.00 / 0.00} & {0.08 / 0.04} & {0.04 / 0.02} & {0.05 / 0.12} & {0.13 / 0.05} & \textcolor{red}{0.52 / 0.05} & {0.01 / 0.01} & {0.08 / 0.03} & {0.04 / 0.01} & {0.00 / 0.05} & \textcolor{blue}{0.02 / 0.35} & {0.02 / 0.00} & {0.00 / 0.00} & {0.02 / 0.00} \\
\hline
prefix\#1 & {0.05 / 0.07} & {0.02 / 0.03} & {0.11 / 0.09} & {0.00 / 0.00} & {0.01 / 0.00} & {0.00 / 0.00} & {0.03 / 0.03} & {0.00 / 0.00} & {0.00 / 0.00} & {0.01 / 0.00} & {0.02 / 0.01} & {0.00 / 0.00} & {0.01 / 0.00} & {0.00 / 0.00} & {0.02 / 0.00} & {0.04 / 0.03} \\
\hline
prefix\#2 & {0.10 / 0.01} & \textcolor{blue}{0.00 / 0.17} & {0.00 / 0.00} & {0.04 / 0.00} & {0.00 / 0.00} & {0.00 / 0.01} & {0.10 / 0.01} & {0.00 / 0.00} & {0.00 / 0.01} & {0.04 / 0.03} & {0.00 / 0.00} & {0.00 / 0.00} & {0.10 / 0.11} & {0.02 / 0.00} & {0.00 / 0.00} & {0.01 / 0.00} \\
\hline
prefix\#3 & {0.00 / 0.00} & {0.01 / 0.03} & \textcolor{red}{0.18 / 0.01} & {0.10 / 0.00} & {0.00 / 0.01} & {0.00 / 0.00} & {0.00 / 0.00} & {0.00 / 0.00} & {0.00 / 0.00} & {0.01 / 0.03} & {0.05 / 0.01} & {0.02 / 0.01} & {0.00 / 0.00} & {0.00 / 0.00} & {0.11 / 0.09} & {0.00 / 0.00} \\
\hline
prefix\#4 & {0.02 / 0.01} & {0.01 / 0.01} & {0.08 / 0.21} & {0.11 / 0.02} & {0.00 / 0.00} & {0.00 / 0.00} & {0.00 / 0.00} & {0.05 / 0.06} & {0.00 / 0.00} & {0.01 / 0.00} & \textcolor{blue}{0.01 / 0.45} & \textcolor{blue}{0.00 / 0.49} & {0.00 / 0.00} & {0.00 / 0.00} & {0.00 / 0.00} & {0.06 / 0.01} \\
\hline
prefix\#5 & {0.00 / 0.00} & {0.13 / 0.04} & {0.00 / 0.00} & {0.00 / 0.00} & {0.00 / 0.00} & {0.00 / 0.00} & {0.03 / 0.01} & {0.02 / 0.04} & {0.00 / 0.00} & \textcolor{blue}{0.00 / 0.21} & {0.08 / 0.25} & {0.01 / 0.03} & {0.00 / 0.02} & {0.03 / 0.01} & {0.01 / 0.00} & {0.00 / 0.01} \\
\hline
prefix\#6 & {0.01 / 0.00} & \textcolor{blue}{0.02 / 0.20} & {0.01 / 0.03} & {0.12 / 0.01} & {0.00 / 0.00} & {0.00 / 0.06} & \textcolor{blue}{0.03 / 0.21} & {0.00 / 0.00} & {0.01 / 0.01} & {0.00 / 0.01} & {0.01 / 0.00} & {0.00 / 0.00} & {0.05 / 0.03} & {0.01 / 0.08} & {0.05 / 0.14} & {0.00 / 0.00} \\
\hline
prefix\#7 & {0.00 / 0.01} & \textcolor{blue}{0.01 / 0.16} & {0.01 / 0.00} & {0.00 / 0.00} & {0.00 / 0.05} & {0.00 / 0.00} & {0.00 / 0.00} & {0.01 / 0.00} & {0.00 / 0.00} & \textcolor{blue}{0.00 / 0.17} & {0.02 / 0.01} & {0.04 / 0.00} & {0.01 / 0.00} & {0.00 / 0.00} & \textcolor{red}{0.22 / 0.01} & {0.00 / 0.00} \\
\hline
prefix\#8 & {0.03 / 0.05} & {0.01 / 0.00} & {0.03 / 0.11} & {0.00 / 0.00} & {0.05 / 0.02} & {0.01 / 0.01} & {0.02 / 0.02} & {0.01 / 0.01} & \textcolor{blue}{0.00 / 0.24} & {0.00 / 0.00} & {0.07 / 0.02} & {0.12 / 0.02} & {0.00 / 0.00} & \textcolor{red}{0.33 / 0.07} & {0.09 / 0.03} & {0.01 / 0.02} \\
\hline
prefix\#9 & {0.00 / 0.00} & {0.01 / 0.01} & {0.00 / 0.00} & {0.00 / 0.00} & {0.00 / 0.00} & {0.00 / 0.00} & {0.01 / 0.01} & {0.00 / 0.00} & {0.00 / 0.00} & {0.00 / 0.00} & {0.00 / 0.07} & {0.00 / 0.00} & {0.00 / 0.00} & {0.00 / 0.00} & {0.06 / 0.00} & {0.01 / 0.01} \\
\hline
text & {0.01 / 0.00} & {0.07 / 0.04} & {0.00 / 0.00} & {0.01 / 0.00} & {0.03 / 0.01} & {0.02 / 0.01} & {0.06 / 0.02} & {0.00 / 0.01} & \textcolor{red}{0.75 / 0.05} & {0.00 / 0.00} & {0.05 / 0.01} & {0.00 / 0.00} & {0.00 / 0.00} & {0.10 / 0.01} & {0.01 / 0.00} & {0.10 / 0.04} \\
\hline
self & {0.44 / 0.39} & {0.09 / 0.04} & {0.44 / 0.35} & {0.06 / 0.20} & {0.30 / 0.23} & {0.03 / 0.12} & {0.02 / 0.03} & {0.24 / 0.58} & {0.01 / 0.01} & {0.21 / 0.24} & {0.03 / 0.00} & {0.04 / 0.05} & {0.30 / 0.19} & {0.13 / 0.10} & {0.11 / 0.11} & {0.05 / 0.02} \\
\hline
context & {0.27 / 0.42} & {0.41 / 0.22} & {0.06 / 0.07} & \textcolor{blue}{0.18 / 0.54} & {0.32 / 0.45} & {0.81 / 0.61} & {0.39 / 0.53} & {0.12 / 0.24} & \textcolor{blue}{0.15 / 0.62} & {0.28 / 0.12} & {0.34 / 0.10} & {0.25 / 0.20} & {0.12 / 0.10} & {0.26 / 0.59} & {0.28 / 0.58} & {0.45 / 0.69} \\
\hline
others & {0.07 / 0.03} & {0.19 / 0.05} & {0.07 / 0.12} & {0.29 / 0.17} & {0.24 / 0.20} & {0.07 / 0.06} & {0.15 / 0.08} & {0.01 / 0.01} & {0.06 / 0.05} & {0.35 / 0.16} & \textcolor{red}{0.27 / 0.06} & \textcolor{red}{0.51 / 0.13} & {0.38 / 0.20} & {0.10 / 0.12} & {0.03 / 0.03} & {0.25 / 0.17} \\
\hline
\end{tabular}
}
\caption{Training-L23}
\end{subtable}

\caption{Encoder attention weights to each sections from all relevant / non-relevant nodes on each head. Values in red are high for relevant nodes, and in blue are high for non-relevant nodes. (The criteria is based on heuristics: $x/y-y/x+x-y>2.0$) Supplementary for Table~\ref{tbl:exp4.1-L23}. The trend is almost the same between training and dev set.
}
\label{tbl:exp4.1-L23-appendix}
\end{table*}

\begin{table}[t]

\begin{subtable}{\linewidth}\centering
\resizebox{\linewidth}{!}{%
\begin{tabular}{|l|c|c|c|}
\hline
Corruption part & Low layers & High layers & All layers \\
\hline
Prefix & 0.9802 & 0.9226 & 0.8499 \\
\hline
Text & 0.9862 & 0.9331 & 0.8594 \\
\hline
Self-node & 0.9625 & 0.8054 & 0.6572 \\
\hline
Struct & 0.9526 & 0.7498 & 0.5507 \\
\hline
Text+struct & 0.9478 & 0.6707 & 0.4909 \\
\hline
StructContext & 0.9876 & 0.9473 & 0.9195 \\
\hline
Text+StructContext & 0.9821 & 0.8757 & 0.6738 \\
\hline
All & 0.9376 & 0.5974 & 0.4509 \\
\hline
\end{tabular}
}
\caption{Columns}
\label{tbl:exp5.2.1-column-appendix}
\end{subtable}

\vspace{0.2cm}

\begin{subtable}{\linewidth}\centering
\resizebox{\linewidth}{!}{%
\begin{tabular}{|l|c|c|c|}
\hline
Corruption part & Low layers & High layers & All layers \\
\hline
Prefix & 0.9796 & 0.8033 & 0.5895 \\
\hline
Text & 0.9913 & 0.9775 & 0.9551 \\
\hline
Self-node & 0.9862 & 0.8126 & 0.4924 \\
\hline
Struct & 0.9631 & 0.7350 & 0.3310 \\
\hline
Text+struct & 0.9545 & 0.6817 & 0.2948 \\
\hline
StructContext & 0.9844 & 0.9384 & 0.7964 \\
\hline
Text+StructContext & 0.9799 & 0.8890 & 0.5685 \\
\hline
All & 0.8988 & 0.4484 & 0.2337 \\
\hline
\end{tabular}
}
\caption{Tables}
\label{tbl:exp5.2.1-table-appendix}
\end{subtable}

\vspace{0.2cm}

\begin{subtable}{\linewidth}\centering
\resizebox{\linewidth}{!}{%
\begin{tabular}{|l|c|c|c|}
\hline
Corruption part & Low layers & High layers & All layers \\
\hline
Prefix & 0.9923 & 0.9541 & 0.9259 \\
\hline
Text & 0.9942 & 0.9856 & 0.9746 \\
\hline
Self-node & 0.9917 & 0.9833 & 0.9519 \\
\hline
Struct & 0.9874 & 0.9468 & 0.9137 \\
\hline
Text+struct & 0.9869 & 0.9405 & 0.9117 \\
\hline
StructContext & 0.9924 & 0.9501 & 0.9384 \\
\hline
Text+StructContext & 0.9911 & 0.9436 & 0.9275 \\
\hline
All & 0.9751 & 0.9278 & 0.9078 \\
\hline
\end{tabular}
}
\caption{Table aliases}
\label{tbl:exp5.2.1-table-alias-appendix}
\end{subtable}

\caption{Attention corruption study on attentions from node-of-interest to each section, the results for all node types. 
}
\label{tbl:exp5.2.1-appendix}
\end{table}

\begin{table*}[t]
\begin{subtable}{\linewidth}\centering
\resizebox{\linewidth}{!}{%

\begin{tabular}{|l|ccc|ccc|}
\hline
\multirow{2}{*}{Corruption part} & \multicolumn{3}{c|}{Weights} & \multicolumn{3}{c|}{Logits} \\
\cline{2-7}
& Low window & High window & All layers & Low layers & High layers & All layers \\
\hline
$T\rightarrow S$ & 0.9908 & 0.9787 & 0.9671 & 0.9836 & 0.9622 & 0.9543 \\\hline
$S\rightarrow T$ & 0.9929 & 0.9647 & 0.9071 & 0.9847 & 0.8381 & 0.6879 \\\hline
$T\leftrightarrow S$ & 0.9889 & 0.9413 & 0.8416 & 0.9762 & 0.7769 & 0.6138 \\\hline
$T\rightarrow P$ & 0.9805 & 0.9460 & 0.8856 & 0.9694 & 0.9251 & 0.8703 \\\hline
$S\rightarrow P$ & 0.9782 & 0.9268 & 0.8459 & 0.9680 & 0.8821 & 0.8317 \\\hline
$TS\rightarrow P$ & 0.9707 & 0.7953 & 0.5793 & 0.9524 & 0.7073 & 0.5778 \\\hline
$T\rightarrow T$ & 0.9882 & 0.9491 & 0.8048 & 0.9578 & 0.9028 & 0.7631 \\\hline
$S\rightarrow S$ & 0.9574 & 0.6447 & 0.5213 & 0.6859 & 0.5708 & 0.4601 \\\hline
all & 0.8502 & 0.3578 & 0.2101 & 0.5213 & 0.2521 & 0.1502 \\\hline
\end{tabular}
}
\caption{Columns}
\label{tbl:exp5.3.*-column-appendix}
\end{subtable}

\vspace{0.2cm}

\begin{subtable}{\linewidth}\centering
\resizebox{\linewidth}{!}{%

\begin{tabular}{|l|ccc|ccc|}
\hline
\multirow{2}{*}{Corruption part} & \multicolumn{3}{c|}{Weights} & \multicolumn{3}{c|}{Logits} \\
\cline{2-7}
& Low window & High window & All layers & Low layers & High layers & All layers \\
\hline
$T\rightarrow S$ & 0.9908 & 0.9784 & 0.9737 & 0.9888 & 0.9774 & 0.9732 \\\hline
$S\rightarrow T$ & 0.9906 & 0.9755 & 0.9319 & 0.9829 & 0.8792 & 0.6925 \\\hline
$T\leftrightarrow S$ & 0.9868 & 0.9654 & 0.9106 & 0.9777 & 0.8498 & 0.6302 \\\hline
$T\rightarrow P$ & 0.9887 & 0.9792 & 0.9471 & 0.9867 & 0.9718 & 0.9469 \\\hline
$S\rightarrow P$ & 0.9695 & 0.7746 & 0.6130 & 0.9642 & 0.7444 & 0.6515 \\\hline
$TS\rightarrow P$ & 0.9625 & 0.7064 & 0.4773 & 0.9547 & 0.6173 & 0.4781 \\\hline
$T\rightarrow T$ & 0.9864 & 0.9833 & 0.9441 & 0.9732 & 0.9802 & 0.9304 \\\hline
$S\rightarrow S$ & 0.9350 & 0.6739 & 0.3604 & 0.5448 & 0.5519 & 0.3727 \\\hline
all & 0.8045 & 0.3939 & 0.1594 & 0.4762 & 0.2360 & 0.1012 \\\hline
\end{tabular}
}
\caption{Tables}
\label{tbl:exp5.3.*-table-appendix}
\end{subtable}



\vspace{0.2cm}

\begin{subtable}{\linewidth}\centering
\resizebox{\linewidth}{!}{%

\begin{tabular}{|l|ccc|ccc|}
\hline
\multirow{2}{*}{Corruption part} & \multicolumn{3}{c|}{Weights} & \multicolumn{3}{c|}{Logits} \\
\cline{2-7}
& Low window & High window & All layers & Low layers & High layers & All layers \\
\hline
$T\rightarrow S$ & 0.9971 & 0.9953 & 0.9933 & 0.9965 & 0.9943 & 0.9919 \\\hline
$S\rightarrow T$ & 0.9980 & 0.9949 & 0.9926 & 0.9977 & 0.9918 & 0.9845 \\\hline
$T\leftrightarrow S$ & 0.9965 & 0.9920 & 0.9878 & 0.9958 & 0.9895 & 0.9788 \\\hline
$T\rightarrow P$ & 0.9945 & 0.9739 & 0.9585 & 0.9947 & 0.9695 & 0.9549 \\\hline
$S\rightarrow P$ & 0.9962 & 0.9853 & 0.9829 & 0.9961 & 0.9857 & 0.9830 \\\hline
$TS\rightarrow P$ & 0.9921 & 0.9459 & 0.8976 & 0.9924 & 0.9401 & 0.9022 \\\hline
$T\rightarrow T$ & 0.9903 & 0.9830 & 0.9432 & 0.9872 & 0.9753 & 0.9268 \\\hline
$S\rightarrow S$ & 0.9834 & 0.9769 & 0.9320 & 0.9674 & 0.9727 & 0.9582 \\\hline
all & 0.9257 & 0.8382 & 0.7269 & 0.8961 & 0.7883 & 0.7338 \\\hline
\end{tabular}
}
\caption{Syntax tokens}
\label{tbl:exp6.1.*-appendix}
\end{subtable}

\caption{Attention corruption study on encoder self-attentions across input sections. For corruption part, "T" means text, "S" for structure, "P" for prefix; "$T\rightarrow S$" means corrupting the attention weights from text to structure tokens, i.e. text tokens as $q$ and structure tokens as $k$.
On top, "Weights" and "Logits" represent the attention corruption scheme.
Supplementary for Table~\ref{tbl:exp5.3+6.1}.}
\label{tbl:exp5.3+6.1-appendix}
\end{table*}

\begin{table*}[t]

\begin{subtable}{\linewidth}\centering
\resizebox{\linewidth}{!}{%
\begin{tabular}{|l|ccc|ccc|}
\hline
\multirow{2}{*}{Corruption part} & \multicolumn{3}{c|}{Weights} & \multicolumn{3}{c|}{Logits} \\
\cline{2-7}
 & Low layers & High layers & All layers & Low layers & High layers & All layers \\
\hline
Text & 0.9588 & 0.9815 & 0.9063 & 0.9565 & 0.9835 & 0.9132 \\
\hline
Struct & 0.8746 & 0.3577 & 0.3239 & 0.8067 & 0.5050 & 0.4333 \\
\hline
Prefix & 0.9260 & 0.9806 & 0.9114 & 0.9199 & 0.9746 & 0.9025 \\
\hline
Struct ctx. & 0.9420 & 0.9867 & 0.9429 & 0.9551 & 0.9981 & 0.9876 \\
\hline
Self-node & 0.9505 & 0.3911 & 0.3840 & 0.9320 & 0.5857 & 0.5382 \\
\hline
All & 0.5805 & 0.1613 & 0.1316 & 0.3284 & 0.1017 & 0.0597 \\
\hline
\end{tabular}}
\caption{Column}
\label{tbl:exp5.4.*-column-appendix}
\end{subtable}

\vspace{0.2cm}

\begin{subtable}{\linewidth}\centering
\resizebox{\linewidth}{!}{%
\begin{tabular}{|l|ccc|ccc|}
\hline
\multirow{2}{*}{Corruption part} & \multicolumn{3}{c|}{Weights} & \multicolumn{3}{c|}{Logits} \\
\cline{2-7}
& Low layers & High layers & All layers & Low layers & High layers & All layers \\
\hline
Text & 0.9895 & 0.9905 & 0.9840 & 0.9907 & 0.9909 & 0.9868 \\
\hline
Struct & 0.7901 & 0.0539 & 0.0672 & 0.7465 & 0.3822 & 0.3137 \\
\hline
Prefix & 0.9273 & 0.9900 & 0.9115 & 0.8963 & 0.9878 & 0.8784 \\
\hline
Struct ctx. & 0.9140 & 0.9940 & 0.9824 & 0.9236 & 0.9997 & 0.9954 \\
\hline
Self-node & 0.9100 & 0.1794 & 0.1738 & 0.9075 & 0.4587 & 0.4272 \\
\hline
All & 0.5965 & 0.0328 & 0.0341 & 0.4779 & 0.0985 & 0.0706 \\
\hline
\end{tabular}}
\caption{Table}
\label{tbl:exp5.4.*-table-appendix}
\end{subtable}

\vspace{0.2cm}


\begin{subtable}{\linewidth}\centering
\resizebox{\linewidth}{!}{%
\begin{tabular}{|l|ccc|ccc|}
\hline
\multirow{2}{*}{Corruption part} & \multicolumn{3}{c|}{Weights} & \multicolumn{3}{c|}{Logits} \\
\cline{2-7}
& Low layers & High layers & All layers & Low layers & High layers & All layers \\
\hline
Text & 0.9820 & 0.9377 & 0.8517 & 0.9770 & 0.9103 & 0.8305 \\\hline
Struct & 0.9568 & 0.9970 & 0.9543 & 0.9317 & 0.9963 & 0.9326 \\\hline
Prefix & 0.9608 & 0.9486 & 0.8332 & 0.9545 & 0.8971 & 0.7676 \\\hline
All & 0.7780 & 0.8878 & 0.4590 & 0.6694 & 0.8221 & 0.3725 \\\hline
\end{tabular}
}
\caption{Syntax tokens}
\label{tbl:exp6.2.*-appendix}
\end{subtable}

\vspace{0.2cm}

\caption{Attention corruption study on decoder cross-attentions to each input section.
Supplementary for Table~\ref{tbl:exp5.4+6.2}.
}
\label{tbl:exp5.4+6.2-appendix}
\end{table*}



\begin{table*}[t]
\centering
\resizebox{\linewidth}{!}{%
\begin{tabular}{|l|ccc|ccc|ccc|}
\hline
\multirow{2}{*}{Text match} & \multicolumn{3}{c|}{Exact (791)} & \multicolumn{3}{c|}{Partial (443)} & \multicolumn{3}{c|}{No match (202)} \\
\cline{2-10}
& Low layers & High layers & All layers & Low layers & High layers & All layers & Low layers & High layers & All layers \\\hline
$T\rightarrow S$ & 0.9942 & 0.9868 & 0.9784 & 0.9820 & 0.9699 & 0.9433 & 0.9882 & 0.9654 & 0.9555 \\\hline
$S\rightarrow T$ & 0.9977 & 0.9838 & 0.9461 & 0.9954 & 0.9525 & 0.8613 & 0.9806 & 0.9289 & 0.8456 \\\hline
$T\leftrightarrow S$ & 0.9954 & 0.9640 & 0.8725 & 0.9839 & 0.9239 & 0.8003 & 0.9773 & 0.8993 & 0.7890 \\\hline
$T\rightarrow P$ & 0.9915 & 0.9601 & 0.9269 & 0.9811 & 0.9346 & 0.8350 & 0.9552 & 0.9183 & 0.8206 \\\hline
$S\rightarrow P$ & 0.9895 & 0.9599 & 0.9307 & 0.9568 & 0.9437 & 0.8441 & 0.9660 & 0.8388 & 0.6529 \\\hline
all & 0.9148 & 0.4417 & 0.2527 & 0.7743 & 0.3159 & 0.2053 & 0.7489 & 0.1492 & 0.0524 \\\hline
\end{tabular}
}
\vspace{0.2cm}

\caption{Attention corruption study on \textbf{encoder self-attentions} across input sections (for column prediction). Comparison of results between target columns with different text-matching situations.
}
\label{tbl:exp5.3.1-col-text-match}
\end{table*}

\begin{table*}[t]
\centering
\resizebox{\linewidth}{!}{%

\begin{tabular}{|l|ccc|ccc|ccc|}
\hline
\multirow{2}{*}{Text match} & \multicolumn{3}{c|}{Exact (791)} & \multicolumn{3}{c|}{Partial (443)} & \multicolumn{3}{c|}{No-match (202)} \\
\cline{2-10}
& Low layers & High layers & All layers & Low layers & High layers & All layers & Low layers & High layers & All layers \\
\hline
Text & 0.9558 & 0.9845 & 0.9006 & 0.9578 & 0.9817 & 0.8859 & 0.9645 & 0.9759 & 0.9252 \\\hline
Struct & 0.9740 & 0.5157 & 0.4410 & 0.7737 & 0.1094 & 0.1353 & 0.7410 & 0.1828 & 0.1946 \\\hline
Prefix & 0.9809 & 0.9896 & 0.9641 & 0.9415 & 0.9744 & 0.9325 & 0.8188 & 0.9670 & 0.8052 \\\hline
StructCtx. & 0.9739 & 0.9924 & 0.9610 & 0.9468 & 0.9572 & 0.8852 & 0.8813 & 0.9901 & 0.9369 \\\hline
Self-node & 0.9926 & 0.5288 & 0.5167 & 0.8994 & 0.2956 & 0.2533 & 0.8979 & 0.1813 & 0.1999 \\\hline
all & 0.7176 & 0.2169 & 0.1877 & 0.4343 & 0.0809 & 0.0672 & 0.3974 & 0.0915 & 0.0530 \\\hline
\end{tabular}
}
\vspace{0.2cm}

\caption{Attention corruption study on \textbf{decoder cross-attentions} to each input section (for column prediction). Comparison of results between target columns with different text-matching situations.
}
\label{tbl:exp5.4-col-text-match}
\end{table*}

\begin{table*}[ht]
\centering
\resizebox{\linewidth}{!}{%
\begin{tabular}{|l|c|c|c|c|c|c|c|c|c||c|c|c|}
\hline
Syntax-tok & $t \rightarrow s$ & $s \rightarrow t$ & $t \leftrightarrow s$ & $t \rightarrow p$ & $s \rightarrow p$ & $ts \rightarrow p$ & $t \rightarrow t$ & $s \rightarrow s$ & \textbf{All} & \textbf{Eff\_cnt} & \textbf{All\_cnt} & \textbf{Eff\_rate}\\
\hline
\texttt{!=} & 0.9984 & 0.9995 & 0.9825 & 0.8485 & 0.9998 & 0.9070 & 0.3025 & 0.9193 & 0.0284 & 20 & 20 & 1.0000 \\
\hline
\texttt{(} & 0.9999 & 0.9794 & 0.9946 & 0.9719 & 0.9795 & 0.9685 & 0.8983 & 0.9467 & 0.1856 & 98 & 675 & 0.1452 \\
\hline
\texttt{)} & 0.9257 & 0.9926 & 0.9716 & 0.9914 & 0.9074 & 0.8420 & 0.9244 & 0.4715 & 0.1108 & 13 & 23 & 0.5652 \\
\hline
\texttt{*} & 1.0000 & 0.9999 & 1.0000 & 0.9997 & 0.9999 & 0.9998 & 0.9999 & 0.9982 & 0.1802 & 10 & 381 & 0.0262 \\
\hline
\texttt{=} & 0.9784 & 0.9923 & 0.9786 & 0.9378 & 0.8940 & 0.7439 & 0.9466 & 0.5933 & 0.1149 & 128 & 968 & 0.1322 \\
\hline
\texttt{>}	& 1.0000	& 1.0000	& 1.0000	& 0.9509	& 0.9990	& 0.9652	& 0.8193	& 0.9845	& 0.0219	& 68	& 101	& 0.6733 \\\hline
\texttt{>=}	& 1.0000	& 1.0000	& 1.0000	& 0.9151	& 1.0000	& 0.9048	& 0.2593	& 0.9964	& 0.1703	& 11	& 30	& 0.3667 \\\hline
and	& 0.7931	& 0.8981	& 0.6727	& 0.4962	& 0.8409	& 0.3468	& 0.3603	& 0.6749	& 0.0554	& 29	& 39	& 0.7436 \\\hline
as	& 0.9212	& 0.8936	& 0.8349	& 0.7992	& 0.5034	& 0.3457	& 0.8478	& 0.3288	& 0.1002	& 52	& 952	& 0.0546 \\\hline
asc	& 0.9723	& 0.9096	& 0.9437	& 0.4939	& 0.8596	& 0.3612	& 0.6133	& 0.6725	& 0.0408	& 19	& 19	& 1.0000 \\\hline
avg	& 0.9996	& 0.9940	& 0.9972	& 0.7660	& 0.9996	& 0.3619	& 0.7743	& 0.9513	& 0.0360	& 63	& 65	& 0.9692 \\\hline
between	& 0.9999	& 0.9999	& 1.0000	& 0.7788	& 0.9999	& 0.3407	& 0.0009	& 0.9999	& 0.0000	& 6	& 6	& 1.0000 \\\hline
count	& 0.9911	& 0.9977	& 0.9960	& 0.9641	& 0.9850	& 0.7469	& 0.8609	& 0.8691	& 0.0220	& 294	& 406	& 0.7241 \\\hline
desc	& 0.9939	& 0.9992	& 0.9942	& 0.9198	& 0.9191	& 0.7296	& 0.9044	& 0.6997	& 0.1144	& 82	& 164	& 0.5000 \\\hline
distinct	& 0.8663	& 0.8651	& 0.8284	& 0.6270	& 0.9118	& 0.4787	& 0.6194	& 0.7636	& 0.0014	& 26	& 26	& 1.0000 \\\hline
except	& 0.9517	& 0.9847	& 0.9415	& 0.6213	& 0.9311	& 0.0783	& 0.4062	& 0.7137	& 0.0002	& 21	& 21	& 1.0000 \\\hline
from	& 0.9924	& 0.9703	& 0.9671	& 0.9781	& 0.9519	& 0.6276	& 0.8407	& 0.5194	& 0.1384	& 263	& 1196	& 0.2199 \\\hline
group	& 0.9895	& 0.9925	& 0.9771	& 0.8332	& 0.9838	& 0.5694	& 0.9167	& 0.7823	& 0.0282	& 241	& 265	& 0.9094 \\\hline
having	& 0.9978	& 0.9924	& 0.9894	& 0.5962	& 0.9993	& 0.3893	& 0.8985	& 0.9348	& 0.0262	& 77	& 81	& 0.9506 \\\hline
in	& 0.9973	& 0.9995	& 0.9992	& 0.9995	& 0.9253	& 0.5859	& 0.7717	& 0.3267	& 0.1096	& 8	& 50	& 0.1600 \\\hline
intersect	& 0.9649	& 0.9938	& 0.9522	& 0.2385	& 0.9592	& 0.0575	& 0.4348	& 0.8243	& 0.0001	& 34	& 34	& 1.0000 \\\hline
join	& 0.9099	& 0.6489	& 0.4976	& 0.7286	& 0.3030	& 0.1984	& 0.7286	& 0.1891	& 0.0158	& 44	& 496	& 0.0887 \\\hline
like	& 1.0000	& 1.0000	& 1.0000	& 0.9997	& 0.9997	& 0.8233	& 0.2700	& 0.8809	& 0.0062	& 12	& 12	& 1.0000 \\\hline
limit	& 0.9932	& 0.9854	& 0.9684	& 0.8400	& 0.9992	& 0.7895	& 0.8076	& 0.7534	& 0.1531	& 26	& 177	& 0.1469 \\\hline
max	& 0.8483	& 0.9925	& 0.9741	& 0.8141	& 0.9458	& 0.4659	& 0.5214	& 0.9008	& 0.0633	& 29	& 30	& 0.9667 \\\hline
min	& 0.9962	& 1.0000	& 1.0000	& 0.9906	& 0.9993	& 0.9377	& 0.7412	& 0.9999	& 0.0450	& 14	& 18	& 0.7778 \\\hline
not	& 0.9884	& 0.9515	& 0.9360	& 0.9510	& 0.9067	& 0.6624	& 0.8928	& 0.7208	& 0.0117	& 45	& 46	& 0.9783 \\\hline
or	& 0.9788	& 0.9728	& 0.9582	& 0.7773	& 0.9856	& 0.9181	& 0.4193	& 0.9718	& 0.0311	& 34	& 34	& 1.0000 \\\hline
order	& 0.9964	& 0.9891	& 0.9826	& 0.7079	& 0.9802	& 0.4836	& 0.8542	& 0.8393	& 0.0680	& 197	& 221	& 0.8914 \\\hline
sum	& 0.9054	& 0.9988	& 0.8969	& 0.8431	& 0.9998	& 0.6812	& 0.4672	& 0.9511	& 0.0119	& 20	& 22	& 0.9091 \\\hline
union	& 0.9990	& 0.8977	& 0.8037	& 0.2034	& 0.7390	& 0.3218	& 0.0174	& 0.7111	& 0.0001	& 6	& 6	& 1.0000 \\\hline
where	& 0.9820	& 0.9900	& 0.9702	& 0.9035	& 0.9576	& 0.7533	& 0.8924	& 0.7699	& 0.0858	& 350	& 516	& 0.6783 \\\hline\end{tabular}
}
\caption{The accuracy of each syntax token when corrupting encoder self-attention.
This table shows the detailed effect of such corruptions.
Eff\_rate means for all occasions with this token as ground truth, the percentage of predictions being affected (became wrong) by the corruption.}
\label{tbl:exp6.1-per-token}
\end{table*}

\begin{table}[t]
\begin{subtable}{1.0\linewidth}
\centering
\resizebox{\linewidth}{!}{%
\begin{tabular}{|c|c|cc|cc|}
\hline
\multirow{2}{*}{Layer} & \multirow{2}{*}{Section} & \multicolumn{2}{c|}{Weights} & \multicolumn{2}{c|}{Logits} \\
\cline{3-6}
& & Exact & Exec & Exact & Exec \\
\hline
Low & $S\rightarrow T$ & 0.6683 & 0.6789 & 0.6538 & 0.6654 \\\hline
Low & $T\rightarrow S$ & 0.6586 & 0.6779 & 0.6518 & 0.6721 \\\hline
Low & $T\leftrightarrow S$ & 0.6538 & 0.6712 & 0.6422 & 0.6634 \\\hline\hline
Mid & $S\rightarrow T$ & 0.6180 & 0.6344 & 0.5019 & 0.5184 \\\hline
Mid & $T\rightarrow S$ & 0.6470 & 0.6692 & 0.6451 & 0.6586 \\\hline
Mid & $T\leftrightarrow S$ & 0.5957 & 0.6141 & 0.4294 & 0.4468 \\\hline\hline
High & $S\rightarrow T$ & 0.5938 & 0.6141 & 0.4072 & 0.4362 \\\hline
High & $T\rightarrow S$ & 0.6248 & 0.6431 & 0.6151 & 0.6412 \\\hline
High & $T\leftrightarrow S$ & 0.5242 & 0.5387 & 0.3250 & 0.3559 \\\hline\hline
all & $S\rightarrow T$ & 0.5145 & 0.5387 & 0.2234 & 0.2369 \\\hline
all & $T\rightarrow S$ & 0.5938 & 0.6190 & 0.5783 & 0.6103 \\\hline
all & $T\leftrightarrow S$ & 0.3926 & 0.4236 & 0.1509 & 0.1712 \\\hline\hline
(clean) & (clean) & 0.6692 & 0.6809 & 0.6692 & 0.6809 \\\hline
\end{tabular}
}
\caption{Encoder self-attention}
\label{tbl:exp-A1.0}
\vspace{0.3cm}
\end{subtable}

\begin{subtable}{1.0\linewidth}
\centering
\resizebox{\linewidth}{!}{%
\begin{tabular}{|c|c|cc|cc|}
\hline
\multirow{2}{*}{Layer} & \multirow{2}{*}{Section} & \multicolumn{2}{c|}{Weights} & \multicolumn{2}{c|}{Logits} \\
\cline{3-6}
& & Exact & Exec & Exact & Exec \\
\hline
Low & Text & 0.5135 & 0.5406 & 0.4400 & 0.4700 \\\hline
Low & Struct & 0.3327 & 0.3472 & 0.2592 & 0.2679 \\\hline\hline
Mid & Text & 0.2737 & 0.2950 & 0.1818 & 0.2002 \\\hline
Mid & Struct & 0.3762 & 0.3801 & 0.2147 & 0.2128 \\\hline\hline
High & Text & 0.4836 & 0.4255 & 0.3994 & 0.3627 \\\hline
High & Struct & 0.0068 & 0.0068 & 0.1015 & 0.1044 \\\hline\hline
all & Text & 0.0996 & 0.1015 & 0.0706 & 0.0812 \\\hline
all & Struct & 0.0010 & 0.0010 & 0.0648 & 0.0638 \\\hline\hline
(clean) & (clean) & 0.6692 & 0.6809 & 0.6692 & 0.6809 \\\hline
\end{tabular}
}
\caption{Decoder cross-attention}
\label{tbl:exp-A1.1}
\vspace{0.3cm}
\end{subtable}

\begin{subtable}{1.0\linewidth}
\centering
\resizebox{\linewidth}{!}{%
\begin{tabular}{|p{5cm}|>{\centering\arraybackslash}p{1.8cm}|>{\centering\arraybackslash}p{1.8cm}|}
\hline
Layer & Exact & Exec \\
\hline
Low & 0.0416 & 0.0329 \\\hline
Mid & 0.0696 & 0.0841 \\\hline
High & 0.2089 & 0.2157 \\\hline
all & 0.0000 & 0.0000 \\
\hline
\end{tabular}
}
\caption{Decoder self-attention. Only experimented with blocking all tokens and using "weights" corruption type, essentially zeroing out the self-attention output vector.}
\label{tbl:exp-A1.2}
\vspace{0.3cm}
\end{subtable}

\vspace{-0.1cm}
\caption{End-to-end SQL performance with different attention corruption settings. Supplementary for Table~\ref{tbl:exp-A1.*}.}
\label{tbl:exp-A1.*-appendix}
\end{table}

\begin{table*}[t]
\centering
\resizebox{\linewidth}{!}{%
\begin{tabular}{|l|p{15cm}|}
\hline
Error class & Definition \\\hline
\multicolumn{2}{|c|}{\textbf{S-class}: Clause-level Semantics Errors} \\\hline
S0  & missing / wrong aggregator  \\\hline
S1  & missing / wrong condition clause or ordering  \\\hline
S2  & missing / wrong literal value  \\\hline
\multicolumn{2}{|c|}{\textbf{N-class}: Node (column / table name) Errors} \\\hline
N0  & invalid (hallucinated) node name, either token or natural language phrases  \\\hline
N1  & using '*' for an actual column  \\\hline
N2  & valid but wrong node name (not including '*')  \\\hline
\multicolumn{2}{|c|}{\textbf{J-class}: Join-chain Errors} \\\hline
J0  & extra join, but still correct  \\\hline
J1  & missing alias reference, may cause ambiguous-column error depending on the schema  \\\hline
\multicolumn{2}{|c|}{\textbf{A-class}: Low-level Syntax Errors} \\\hline
A0  & Unpaired brackets / quotes  \\\hline
A1  & Misspelled keyword  \\\hline
A2  & Non-ending token  \\\hline
\multicolumn{2}{|c|}{\textbf{B-class}: Clause-level Syntax Errors} \\\hline
B0  & Missing / extra / misplaced / partial clauses (causing syntax error)  \\\hline
B2  & Alias error (t1 -> t1.col, t1 -> t1st, errors like this)  \\\hline
B3  & Missing / extra operator (causing syntax error)  \\\hline
\multicolumn{2}{|c|}{\textbf{C-class}: Other High-level Semantics Errors} \\\hline
C0  & Natural language expression of SQL, or unquoted string values  \\\hline
C3  & (Not really an error) - equivalent or alternative correct answer  \\\hline
\end{tabular}
}
\caption{Decoder cross-attention corruption error analysis: detailed categories and descriptions.}
\label{tbl:exp-A1.1.1-error-analysis-criteria}
\vspace{0.3cm}
\end{table*}

\begin{table*}[t]
\centering
\resizebox{\linewidth}{!}{%
\begin{tabular}{|c|c|c|c|c|c|c|c|c|c|c|c|c|c|c|c|c|}
\hline
\multirow{2}{*}{\textbf{Category}} & \multicolumn{3}{c|}{\textbf{S}} & \multicolumn{2}{c|}{\textbf{J}} & \multicolumn{3}{c|}{\textbf{N}} & \multicolumn{3}{c|}{\textbf{A}} & \multicolumn{3}{c|}{\textbf{B}} & \multicolumn{2}{c|}{\textbf{C}} \\
\cline{2-17}
 & \textbf{S0} & \textbf{S1} & \textbf{S2} & \textbf{J0} & \textbf{J1} & \textbf{N0} & \textbf{N1} & \textbf{N2} & \textbf{A0} & \textbf{A1} & \textbf{A2} & \textbf{B0} & \textbf{B2} & \textbf{B3} & \textbf{C0} & \textbf{C3} \\
\hline
all-text & 31 & 21 & 1 & 0 & 1 & 0 & 6 & 2 & 0 & 0 & 0 & 0 & 0 & 0 & 0 & 1 \\
\hline
all-struct & 0 & 0 & 0 & 0 & 0 & 46 & 0 & 1 & 1 & 1 & 1 & 6 & 1 & 2 & 1 & 0 \\
\hline
\end{tabular}
}
\caption{Decoder cross-attention corruption error analysis: detailed break down.}
\label{tbl:exp-A1.1.1-error-analysis-break-down}
\vspace{0.3cm}
\end{table*}

\begin{table*}[t]
\centering
\resizebox{\linewidth}{!}{%
\begin{tabular}{|l|p{15cm}|}
\hline
Error class & Definition \\\hline
\multicolumn{2}{|c|}{\textbf{A-class}: Low-level Syntax Errors} \\\hline
A0	& Unpaired brackets / quotes \\\hline
A1	& Misspelled keyword (some keywords like `avg', `distinct' could be tokenized)\\\hline
A2	& Non-ending token \\\hline
\multicolumn{2}{|c|}{\textbf{B-class}: Clause-level Syntax Errors} \\\hline
B0	& Missing / extra / misplaced clauses (causing syntax error) \\\hline
B2	& Alias error (t1 -> t1.col, t1 -> t1st, errors like these) \\\hline
B3	& Missing / extra operator \\\hline
\multicolumn{2}{|c|}{\textbf{C-class}: High-level Semantics Errors} \\\hline
C0	& Natural language expression of SQL, or unquoted string values \\\hline
C1	& Wrong node name with similar surface form \\\hline
C2	& Valid SQL but wrong semantics from user query \\\hline
C3	& (Not really an error) - equivalent or alternative correct answer \\\hline
\end{tabular}
}
\caption{Decoder self-attention corruption error analysis: detailed categories and descriptions.}
\label{tbl:exp-A1.2.0-error-analysis-criteria}
\vspace{0.3cm}
\end{table*}

\begin{table*}[t]
\centering
\resizebox{\linewidth}{!}{%
\begin{tabular}{|l|c|c|c|c|c|c|c|c|c|c|c|}
\hline
\multirow{2}{*}{\textbf{Category}} & \multicolumn{3}{c|}{\textbf{A}} & \multicolumn{4}{c|}{\textbf{B}} & \multicolumn{4}{c|}{\textbf{C}} \\
\cline{2-12}
 & \textbf{A0} & \textbf{A1} & \textbf{A2} & \textbf{B0} & \textbf{B1} & \textbf{B2} & \textbf{B3} & \textbf{C0} & \textbf{C1} & \textbf{C2} & \textbf{C3} \\
\hline
low & 31 & 19 & 5 & 10 & 1 & 4 & 0 & 0 & 0 & 1 & 0 \\
\hline
mid & 27 & 0 & 0 & 16 & 0 & 0 & 3 & 1 & 0 & 1 & 2 \\
\hline
high & 0 & 0 & 0 & 14 & 0 & 1 & 9 & 21 & 4 & 1 & 2 \\
\hline
\end{tabular}
}
\caption{Decoder self-attention corruption error analysis: detailed break down.}
\label{tbl:exp-A1.2.0-error-analysis-break-down}
\vspace{0.3cm}
\end{table*}



\end{document}